\documentclass[onefignum,onetabnum]{siamart190516}
\usepackage{amsmath,amsfonts,amssymb}
\usepackage{graphicx}
\usepackage{xcolor}
\usepackage{booktabs}
\usepackage{multirow}

\usepackage{tikz}
\usetikzlibrary{circuits.ee.IEC, positioning, shapes, arrows}
\tikzstyle{bus}=[rectangle, inner sep=0pt, minimum height=0pt, minimum width=30pt, draw]

\DeclareMathOperator*{\argmax}{arg\,max}

\title{Physics-Informed Gaussian Process Regression for Probabilistic States Estimation and Forecasting in Power Grids
\thanks{ 
    \funding{This work was partially supported by the U.S. Department of Energy (DOE) Office of Science, Office of Advanced Scientific Computing Research (ASCR) as part of the Multifaceted Mathematics for Rare, Extreme Events in Complex Energy and Environment Systems (MACSER) project. Pacific Northwest National Laboratory is operated by Battelle for the DOE under Contract DE-AC05-76RL01830.
}}}

\author{Tong Ma\thanks{Pacific Northwest National Laboratory (\email{tong.ma@pnnl.gov}, \email{david.barajas-solano@pnnl.gov}, \email{ramakrishna.tipireddy@pnnl.gov})}
  \and David Alonso Barajas-Solano\footnotemark[2]
  \and Ramakrishna Tipireddy\footnotemark[2]
  \and Alexandre Tartakovsky\thanks{University of Illinois at Urbana-Champaign and Pacific Northwest National Laboratory (\email{amt1998@illinois.edu})}
  }

\begin{document}

\maketitle

\begin{abstract}
  Real-time state estimation and forecasting is critical for efficient operation of power grids. 
  In this paper, a physics-informed Gaussian process regression (PhI-GPR) method is presented and used for probabilistic forecasting and estimating the phase angle, angular speed, and wind mechanical power of a three-generator power grid system using sparse measurements.
  In standard data-driven Gaussian process regression (GPR), parameterized models for the prior statistics are fit by maximizing the marginal likelihood of observed data, whereas in PhI-GPR, we compute the prior statistics by solving stochastic equations governing power grid dynamics.
  The short-term forecast of a power grid system dominated by wind generation is complicated by the stochastic nature of the wind and the resulting uncertain  mechanical wind power.
  Here, we assume that the power-grid dynamic is governed by the swing equations, and we treat the unknown terms in the swing equations (specifically, the mechanical  wind power) as random processes, which turns these equations into stochastic differential equations.
  We solve these equations for the mean and variance of the power grid system using the Monte Carlo simulations method.
  
  We demonstrate that the proposed PhI-GPR method can accurately forecast and estimate both observed and unobserved states, including the mean behavior and associated uncertainty.
  For observed states, we show that PhI-GPR provides a forecast comparable to the standard data-driven GPR, with both forecasts being significantly more accurate than the  autoregressive integrated moving average (ARIMA) forecast.  
  We also show that the ARIMA forecast is much more sensitive to observation frequency and measurement errors than the PhI-GPR forecast.
\end{abstract}


\section{Introduction}

Real-time monitoring and short-term forecasting of power grid states are important for the grid's control and planning, including power flow optimization and anomaly detection.
Other applications requiring real-time monitoring and short-term forecasting include efficient operation of controllers and the determination of necessary corrective actions against possible failures in power grid systems~\cite{huang2012state}.
Although modern power grids are heavily instrumented, it still remains a challenge to measure all the power grid states due to the inherent high-frequency oscillations of power grid dynamics and increasing penetration of renewable energy sources.
Hence, it is necessary to develop new algorithms for immediate probabilistic forecasting of observed and unobserved states (including the mean behavior and the associated uncertainty) such that power grid systems can operate with efficiency and safety. 

Here, we distinguish between forecasting (extrapolation) and state estimation, which, for the purpose of this work, we define as computing unobserved states in the (recent) past from the values of observed states. %
There are two general types of forecasting methods, including machine learning (ML) forecasting techniques (e.g., fuzzy regression models~\cite{song2005short,hong2014fuzzy}, support vector machines~\cite{fentis2016short}, deep neural networks~\cite{grant2014short,yeung2017learning,thiyagarajan2016real}, gradient boosting machines~\cite{taieb2014gradient,lloyd2014gefcom2012}, Gaussian process regression (GPR)~\cite{williams2006gaussian,quinonero2005unifying,snelson2006sparse}), and statistical techniques (e.g., Markov chains~\cite{yoder2014short}, data mining~\cite{kusiak2009short}, multiple linear regression models~\cite{hong2010short,hong2013long,charlton2014refined,wang2016electric}, semi-parametric additive models~\cite{hyndman2009density,fan2011short,goude2013local,nedellec2014gefcom2012}, autoregressive integrated moving average (ARIMA) models~\cite{hyndman2018forecasting,brockwell2016introduction}, and exponential smoothing models~\cite{hyndman2018forecasting,hong2014global}). There are also various hybrid
approaches that combine some elements of machine learning and statistics~\cite {catalao2011hybrid,liu2012short}.
In general, all ML and statistics methods work better for interpolation than extrapolation. Forecasting is an extrapolation problem and, therefore, there exist no perfect ML or statistical forecasting models.
Most importantly, ML methods cannot estimate or forecast unobserved states because ML methods require observations for training. 

If fully known, physics-based models should be able to accurately forecast the dynamics of complex systems because the conservation laws these models are based on hold both in the past and future.
However, for complex systems such as power grid systems, physics-based models (e.g., models based on the swing equations) are not fully known.
For example, in a model of an electrical grid dominated by wind energy, the mechanical wind power is uncertain, and no amount of observations can predict the mechanical wind power in the future with absolute certainty. 
Therefore, physics-based models alone cannot be used for accurately forecasting the states of the power grid. 

In this work, we present the physics-informed GPR (PhI-GPR) method for dynamic systems governed by a system of ordinary differential equations (ODEs).
In GPR, a power grid state is represented as a linear combination of measured values of said state, with the coefficients being a function of the prior mean and covariance function of the state.
In the standard ``data-driven'' GPR, prior statistics is chosen by fitting parameterized models.
The hyperparameters of these models are found by maximizing the marginal likelihood of the observations~\cite{williams2006gaussian}.
 In PhI-GPR, we assume that  power-grid dynamics is governed by the swing equations, and we treat the unknown mechanical wind power in these equations as a random process, which turns these equations into stochastic differential equations.
We solve these equations for the mean and covariance of the power grid's state using the Monte Carlo simulations method.

The main idea behind PhI-GPR (of computing a prior statistics of a stochastic dynamic system from a partially known stochastic physics-based model of this system) was originally proposed in \cite{tartakovsky2019physicsHICSS} for a simple system of equations describing a single wind-powered generator. Here, we formulate PhI-GPR for an arbitrary large system of ODEs and provide a critical comparison with data-driven GPR and ARIMA forecasting methods.  While the proposed Phi-GPR method is applicable to (among other dynamic systems) a power grid with an arbitrary large number of generators, here we use it to model a power grid with two wind generators and one traditional generator.  
We demonstrate that for the considered power grids with states oscillating around equilibrium due to random wind variations, the PhI-GPR method is able to accurately estimate and forecast unobserved states with the same accuracy as observed states.
For observed states, we find that the accuracy of a PhI-GPR forecast is comparable to the accuracy of a ``data-driven'' GPR forecast.
For forecasting of observed variables, we also compare the PhI-GPR method against the ARIMA method, a commonly used forecasting technique.
We find that ARIMA is sensitive to observation frequency and measurement noise.
We show that PhI-GPR and ARIMA have a comparable accuracy for forecasting observed states with noiseless observations and a short time between observations.
Nevertheless, in the presence of noise and/or a large time interval between observations, the accuracy of ARIMA deteriorates faster than that of PhI-GPR. 
Most importantly, the PhI-GPR method can forecast and estimate both observed and unobserved states, while data-driven GPR and ARIMA can only be used for forecasting and estimating observed states because the observation data are needed to ``train'' these methods.

This paper is organized as follows.
In Section~\ref{sec:phi-GPR}, we introduce the PhI-GPR method, and in Section~\ref{sec:phi-gpr-power-grid} we introduce the stochastic model of the power grid.
In Section~\ref{sec:results}, we apply the PhI-GPR method to the forecasting and state estimation of a synthetic system.
We also compare PhI-GPR forecasts with standard data-driven GPR and ARIMA forecasts.
Conclusions are presented in Section~\ref{sec:Conclusions}.

\section{PhI-GPR method}
\label{sec:phi-GPR}

In this section, we present the PhI-GPR method for forecasting observed and unobserved states of dynamical systems.
The formulation of multivariate GPR for forecasting is described in Section~\ref{sec:multivariate-GPR}.
In Section~\ref{sec:phi-gpr-priors}, we describe how the physics-based priors for multivariate GPR and PhI-GPR are evaluated.

\subsection{Multivariate GPR for forecasting observed and unobserved states}
\label{sec:multivariate-GPR}

We assume that the system is composed of $N$ observed states, $x_i(t)$ ($i = 1, \dots, N$), and $M$ unobserved states $y_i(t)$, ($i = 1, \dots, M$), and that there are measurements $x_{i, j}$ of the observed states at the $N_{t^o}$ discrete times $t^o_j$ $(j = 1, \dots, N_{t^o})$ over the observation window $[0, T^o]$, contaminated by observation noise.
We assume that the observation errors are normal, identically distributed, and uncorrelated across times and states; therefore, we model the observations as
\begin{equation*}
  x_{i, j} = x_i(t^o_j) + \epsilon_{i, j}, \quad \epsilon_{i, j} \sim \mathcal{N}(0, \sigma_n),
\end{equation*}
where $\sigma_n$ is the standard deviation of the observation errors $\epsilon_{i, j}$.
We organize these observations into the vector
\begin{equation*}
  X^o = [x_{1, 1}, x_{1, 2}, \dots, x_{1, N_{t^o}}, \dots x_{N, 1}, x_{N, 2}, \dots, x_{N, N_{t^o}}]^{\top}.
\end{equation*}

First, we are interested in forecasting the observed states for the discrete times $t^f_j$ $(j = 1, \dots, N_{t^f})$ over the forecast window $(T^o, T^f]$; that is, we want to estimate the vector of values
\begin{equation*}
  X^f = [x_1(t^f_1), x_1(t^f_2), \dots, x_1(t^f_{N_{t^f}}), \dots, x_N(t^f_1), x_N(t^f_2), \dots, x_N(t^f_{N_{t^f}})]^{\top}.
\end{equation*}
For this purpose, we employ multivariate GPR regression, in which we model the vectors $X^o$ and $X^f$ as realizations of the random vector $X^{\top} = [(X^o)^{\top}, (X^f)^{\top}]$ with distribution,
\begin{equation}
  \label{eq:gp}
  \begin{bmatrix}
    X^o \\ X^f
  \end{bmatrix} \sim \mathcal{N} \left (
    \begin{bmatrix}
      \bar{X}^o \\ \bar{X}^{f}
    \end{bmatrix},
    \begin{bmatrix}
      K_{X^o X^o} & K_{X^o X^f} \\
      K_{X^o X^f}^T &  K_{X^f X^f}
    \end{bmatrix}
  \right ),
\end{equation}
where $\bar{X}^o$ and $\bar{X}^{f}$ are the so-called prior (or unconditional) mean vector of $X^o$ and $X^f$, respectively, and $K_{X^o X^o}$, $K_{X^o X^f}$, and $K_{X^f X^f}$ are the prior covariance matrices between $X^o$ and $X^o$, $X^o$ and $X^f$, and $X^f$ and $X^f$, respectively.
The covariance matrix $K_{X^o X^f}$ has the block structure
\begin{equation}
  \label{eq:Kof}
  K_{X^o X^f} =
  \begin{bmatrix}
    K_{x_1, x_1}(t^o, t^f) & K_{x_1, x_2}(t^o, t^f) & \cdots & K_{x_1, x_N}(t^o, t^f)\\
    K_{x_2, x_1}(t^o, t^f) & K_{x_2, x_2}(t^o, t^f) & \cdots & K_{x_2, x_N}(t^o, t^f)\\
    \vdots & \vdots & \ddots & \vdots\\
    K_{x_N, x_1}(t^o, t^f) & K_{x_N, x_2}(t^o, t^f) & \cdots & K_{x_N, x_N}(t^o, t^f)\\
  \end{bmatrix},
\end{equation}
where each component $K_{x_i, x_j}(t^o, t^f)$ is given by
\begin{equation}
  \label{eq:Kij-ab}
  K_{\alpha, \beta}(t^a, t^b) =
  \begin{bmatrix}
    \langle \alpha(t^a_1) \beta(t^b_1) \rangle & \cdots & \langle \alpha(t^a_1) \beta(t^b_{N_{t^b}}) \rangle\\
    \vdots & \ddots & \vdots\\
    \langle \alpha(t^a_{N_{t^a}}) \beta(t^b_1) \rangle & \cdots & \langle \alpha(t^a_{N_{t^a}}) \beta(t^b_{N_{t^b}}) \rangle\\
  \end{bmatrix},
\end{equation}
with $\alpha = x_i$, $\beta = x_j$, $a = o$, and $b = f$, and where $\langle \cdot \rangle$ denotes the expectation operator.
The covariance matrix $K_{X^f X^f}$ has a similar structure, with blocks given by \eqref{eq:Kij-ab} with $\alpha = x_i$, $\beta = x_j$, $a = f$, and $b =f$.
Finally, the covariance matrix $K_{X^o X^o}$ has the structure
\begin{equation}
  \label{eq:Koo}
  K_{X^o X^o} =
  \begin{bmatrix}
    K_{x_1, x_1}(t^o, t^o) & K_{x_1, x_2}(t^o, t^o) & \cdots & K_{x_1, x_N}(t^o, t^o)\\
    K_{x_2, x_1}(t^o, t^o) & K_{x_2, x_2}(t^o, t^o) & \cdots & K_{x_2, x_N}(t^o, t^o)\\
    \vdots & \vdots & \ddots & \vdots\\
    K_{x_N, x_1}(t^o, t^o) & K_{x_N, x_2}(t^o, t^o) & \cdots & K_{x_N, x_N}(t^o, t^o)\\
  \end{bmatrix} + \sigma^2_n I
\end{equation}
where $I$ denotes the $(N_{t^o} N) \times (N_{t^o} N)$ identity matrix, and $\sigma_n$ is again the standard deviation of the observation noise.
The addition of the term $\sigma^2_n I$ accounts for observation noise.

Given the state observations, the prior mean vectors, and the covariance matrices, the conditional (or posterior) estimate of the forecast vector $X^{f}$ is given by
\begin{equation}
  \label{eq:sef}
  \hat{X}^{f} = \bar{X}^f + K_{X^o X^f}^T K_{X^o X^o}^{-1} \left(X^o - \bar{X}^o \right),
\end{equation}
and the posterior covariance is given by
\begin{equation}
  \label{eq:cef}
  \hat{K}_{X^f X^f} = K_{X^f X^f} - K_{X^o X^f}^T K_{X^o X^o}^{-1} K_{X^o X^f}.
\end{equation}
The posterior covariance provides a measure of uncertainty or credibility for the forecast of~\eqref{eq:sef}.

We now consider the forecasting of the unobserved states $y_i(t)$.
Our goal is to estimate these states over the observation window $[0, T^o]$ and to forecast them over the forecast window $(T^o, T^f]$.
For simplicity, we consider the forecasting of unobserved states at the discrete times $t^f_j$, $j = 1, \dots, N_{t^f}$ over the time window $(T^o, T^f]$, which we perform again using multivariate GPR.
For this purpose we introduce the vector
\begin{equation*}
  Y^f = [y_1(t^f_1), y_1(t^f_2), \dots, y_1(t^f_{N_{t^f}}), \dots, y_N(t^f_1), y_N(t^f_2), \dots, y_N(t^f_{N_{t^f}})]^{\top},
\end{equation*}
and the random vector $X^{\top} = [(X^o)^{\top}, (Y^f)^{\top}]$ with distribution,
\begin{equation}
  \label{eq:gp-unobserved}
  \begin{bmatrix}
    X^o \\ Y^f
  \end{bmatrix} \sim \mathcal{N} \left (
    \begin{bmatrix}
      \bar{X}^o \\ \bar{Y}^{f}
    \end{bmatrix},
    \begin{bmatrix}
      K_{X^o X^o} & K_{X^o Y^f} \\
      K_{X^o Y^f}^T &  K_{Y^f Y^f}
    \end{bmatrix}
  \right ),
\end{equation}
where $\bar{X}^o$ and $\bar{Y}^{f}$ are the prior mean vectors of $X^o$ and $Y^f$, respectively, and $K_{oo}$, $K_{of}$, and $K_{ff}$ are the prior covariance matrices between $X^o$ and $X^o$, $X^o$ and $Y^f$, and $Y^f$ and $Y^f$, respectively.
Here, $K_{X^o X^o}$ is given by~\eqref{eq:Koo}.
The remaining covariances, $K_{X^o Y^f}$ and $K_{Y^f Y^f}$, have again the structure of \eqref{eq:Kof}, but with blocks $K_{x_i, y_j}(t^o, t^f)$ and $K_{y_i, y_j}(t^f, t^f)$ given by~\eqref{eq:Kij-ab}.

\begin{align*}
  \hat{Y}^{f} &= \bar{Y}^f + K_{X^o Y^f}^T K_{X^o X^o}^{-1} \left (X^o - \bar{X}^o \right),\\
  \hat{K}_{Y^f Y^f} &= K_{Y^f Y^f} - K_{X^o Y^f}^T K_{X^o X^o}^{-1} K_{X^o Y^f}.
\end{align*}

We note that multivariate GPR can be used to estimate ``missing'' observations in incomplete time series, which is a standard regression or interpolation task.
This can be accomplished by adding the time of the missing observation $t^e_j \in [0, T^o]$ $(j = 1, \dots, N_{t^e})$ to the vector of forecast values $X^f$ defined above, where $N_{t^e}$ is the number of missing observation times to be estimated.

\subsection{Prior statistics for the PhI-GPR method}
\label{sec:phi-gpr-priors}

Selecting prior statistics (prior mean and covariance) for the multivariate GP model~\eqref{eq:gp} is one of the main challenges in GPR.
In standard data-driven GPR, prior statistics are often selected from parametric models by maximizing the so-called marginal likelihood of the observations $X^o$~\cite{williams2006gaussian}.
For the case of forecasting observed states, this approach consists of assuming  parametric models for $\bar{X}^o$, $\bar{X}^f$, and the block components $K_{x_i, x_j}(t^a, t^b)$ of $K_{X^o X^o}$, $K_{X^o X^f}$, and $K_{X^f X^f}$, with hyperparameters $\gamma$.
A point estimate of these hyperparameters $\gamma^{*}$ is computed as the value that maximizes the marginal likelihood of the observations, that is,
\begin{equation}
  \label{eq:marginal-likelihood}
  \gamma^{*} = \argmax_{\gamma} \left \{ - \frac{1}{2} (X^o)^{\top} K^{-1}_{X^o X^o} (X^o) - \frac{1}{2} \log \det K_{X^o X^o} - \frac{N_{t^o}}{2} \log 2 \pi \right \}
\end{equation}
while ensuring that the covariance of the joint process $X$ is positive definite.
For a review of prior models for multivariate GP regression, see~\cite{genton-2015-crosscovariance}.

This data-driven GPR approach cannot be employed when there are no observations available for a subset of states to be forecasted.
This is because the marginal likelihood of \eqref{eq:marginal-likelihood} does not include block terms of the form $K_{x_i y_j}(\cdot, \cdot)$ or $K_{y_i y_j}(\cdot, \cdot)$.
Therefore, the covariance of unobserved states and the cross-covariance between observed and unobserved states cannot be estimated by marginal likelihood maximization.

To address this challenge, in PhI-GPR we assume that the observations $X^o$ correspond to noisy observations of the system described by the stochastic process $x_i(t; \omega)$ $(i = 1, \dots, N)$ and $y_i(t; \omega)$ $(i = 1, \dots, M)$, and that we count with a stochastic model (such as a stochastic differential equation, stochastic difference equation, etc.) that can be used to simulate realizations of the stochastic processes $x_i(t; \omega)$ and $y_i(t; \omega)$.

In PhI-GPR, we employ this stochastic model to compute the prior statistics for multivariate GPR via simple Monte Carlo simulations.
For the multivariate GP models~\eqref{eq:gp} and~\eqref{eq:gp-unobserved}, the simple Monte Carlo estimates of $\bar{X}^o$, $\bar{X}^f$, and $\bar{Y}^f$ are given by
\begin{gather*}
  \bar{X}^o = [ \bar{x}_{1} (t^o_{1}), \bar{x}_{1} (t^o_{2}), \dots, \bar{x}_{1} (t^o_{N_{t^o}}), \dots, \bar{x}_{N} (t^o_{1}), \bar{x}_{N} (t^o_{2}), \dots, \bar{x}_N (t^o_{N_{t^o}}) ]^{\top},\\
  \bar{X}^f = [ \bar{x}_{1} (t^f_{1}), \bar{x}_{1} (t^f_{2}), \dots, \bar{x}_{1} (t^f_{N_{t^f}}), \dots, \bar{x}_{N} (t^f_{1}), \bar{x}_{N} (t^f_{2}), \dots, \bar{x}_N (t^f_{N_{t^f}}) ]^{\top},\\
  \bar{Y}^f = [ \bar{y}_1(t^f_1), \bar{y}_1(t^f_2), \dots, \bar{y}_1(t^f_{N_{t^f}}), \dots, \bar{y}_N(t^f_1), \bar{y}_N(t^f_2), \dots, \bar{y}_N(t^f_{N_{t^f}})]^{\top}.
\end{gather*}
Here, each component is given by the simple Monte Carlo estimate
\begin{equation*}
  \bar{\alpha}(t) = \frac{1}{N_{\mathrm{MC}}} \sum^{N_{\mathrm{MC}}}_{n = 1} \alpha (t; \omega^{(n)} ),
\end{equation*}
where $\alpha(t, \omega^{(n)})$ denotes the $n^{\text{th}}$ simulated realization of the state $\alpha$, and $N_{\mathrm{MC}}$ is the number of random simulations of the stochastic model.
Similarly, the covariance components of the covariance matrices in the multivariate GP models~\eqref{eq:gp} and~\eqref{eq:gp-unobserved} are estimated using the simple Monte Carlo estimate
\begin{equation*}
  \langle \alpha(t) \beta(\tau) \rangle = \frac{1}{N_{\mathrm{MC}} - 1} \sum^{N_{\mathrm{MC}}}_{n = 1} (\alpha(t; \omega^{(n)})-\bar{\alpha}(t)) (\beta(\tau; \omega^{(n)} )-\bar{\beta}(\tau)).
\end{equation*}

\section{Power grid model}
\label{sec:phi-gpr-power-grid}

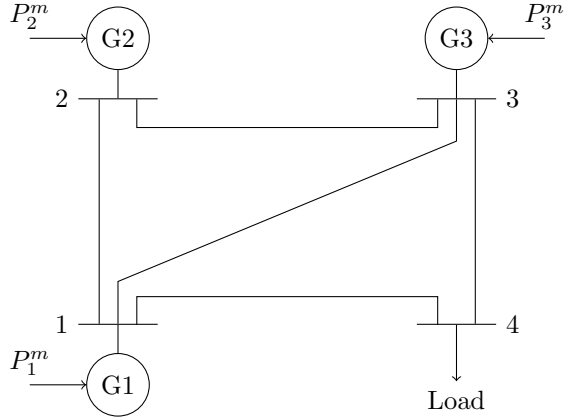
\begin{figure}[tpbh]
  \centering
  \begin{tikzpicture}[circuit ee IEC, x = 0.75cm, y = 0.75cm]
    \node [bus, label=left:$1$]  (b1) at (0, 1) {}; %
    \node [bus, label=left:$2$]  (b2) at (0, 5) {}; %
    \node [bus, label=right:$4$] (b3) at (6, 1) {}; %
    \node [bus, label=right:$3$] (b4) at (6, 5) {}; %
    \node [draw, circle] (g1) [below=0.5 of b1.south] {$\mathrm{G}1$}; %
    \node [draw, circle] (g2) [above=0.5 of b2.north] {$\mathrm{G}2$}; %
    \node [draw, circle] (g4) [above=0.5 of b4.north] {$\mathrm{G}3$}; %
    \draw (g1.north) -- (b1.south); %
    \draw (g2.south) -- (b2.north); %
    \draw (g4.south) -- (b4.north); %
    \path [draw] ([xshift=-0.25cm]b1.north) -- ([xshift=-0.25cm]b2.south); %
    \path [draw] ([xshift=0.25cm]b3.north) -- ([xshift=0.25cm]b4.south); %
    \path [draw] ([xshift=0.25cm]b2.south) --++ (0.0, -0.5) -| ([xshift=-0.25cm]b4.south); %
    \path [draw] ([xshift=0.25cm]b1.south) --++ (0.0, 0.5) -| ([xshift=-0.25cm]b3.south); %
    \path [draw] (b1.north) --++ (0.0, 0.75) -- (6.0, 4.25) -- (b4.south); %
    \draw [->] (b3.south) --++(0.0, -1.0) node [below] {Load}; %
    \draw [<-] (g1.west) --++(-1.0, 0.0) node [above] {$P^m_1$}; %
    \draw [<-] (g2.west) --++(-1.0, 0.0) node [above] {$P^m_2$}; %
    \draw [<-] (g4.east) --++(1.0, 0.0) node [above] {$P^m_3$}; %
  \end{tikzpicture}
  \caption{Schematic of a power system composed of three synchronous generators and four buses: 3 generator buses and 1 load bus.
    Wind mechanical power driving synchronous generators indicated by $P^m_k$, $k \in [1, 3]$.}
  \label{fig:powergrid}
\end{figure}

We consider a power transmission network, with power generators modeled as classical generators driven by mechanical wind power.
We assume that the dynamics of the system can be fully described by swing equations for each generator in the network, together with a constant impedance model for the loads~\cite{nishikawa2015comparative}.
Furthermore, we assume that mechanical wind power is not known deterministically and follows a Langevin equation.

In this work we consider the dynamics of the network shown in Figure~\ref{fig:powergrid}, consisting of three classical generators driven by mechanical wind power and one load, and described by the following equations~\cite{wang2015probabilistic}:
\begin{align}
  \label{eq:powergrid_eq_theta}
  {{\dot \theta }_k} &= {\omega _B}({\omega _k} - {\omega _s}),\\
  \label{eq:powergrid_eq_omega}
  2{H_k}{{\dot \omega }_k} &=  - {D_k}({\omega _k} - {\omega _s}) - P_k^e({\bf{\theta }}) + P_{k}^m, \quad k = 1,2,3,\\
  \label{eq:powergrid_eq_pf}
  P_k^e({\bf{\theta }}) &= \sum\limits_{i = 1}^N {{E_k}} {E_i}\left( {{G_{ki}}\cos ({\theta _k} - {\theta _i}) + {B_{ki}}\sin ({\theta _k} - {\theta _i})} \right).
\end{align}
Here, $\omega_k$ and $\theta_k$ are the angular velocity [rad $\mathrm{s}^{-1}$] and angle [rad] of the $k^\text{th}$ generator, $H_k$ [s] and $D_k$ [p.u.] are the generators' inertia and damping constants, $\omega_B$ [rad $\mathrm{s}^{-1}$] is the base speed, $\omega_s$ [rad $\mathrm{s}^{-1}$] is the synchronization speed, $E_k$ is the phasor internal electromotive force (emf) of the $k^\text{th}$ generator [p.u.], and $G_{ki}$ [p.u.] and $B_{ki}$ [p.u.], $k, i \in [1, 3]$, are the transfer conductances and susceptances, respectively.
Finally, $P_k^e$ [p.u.] and $P_{k}^m$ [p.u.] are the active generated power and mechanical wind power injection of the $k^\text{th}$ generator, respectively.

Unless mentioned otherwise, we assume that we have frequent (every 0.05 s) measurements of $\omega_k$ and $\theta_k$ and that no measurements or an accurate deterministic model of $P^m_k$ are available.
Given the available observations, we are interested in forecasting the grid's dynamics subject to the initial conditions $\theta_k(0) = \theta_{k,0}$ and $\omega_k(0) = \omega_{k,0}$, $k \in [1, 3]$.
Because $P_{k}^m(t)$ is an unknown function of time, without additional data or assumptions about  wind power, equations~\eqref{eq:powergrid_eq_theta}-\eqref{eq:powergrid_eq_pf} cannot be used to forecast power grid states.

In PhI-GPR, we treat $P_{k}^m(t)$ as a stochastic process, which turns equations~\eqref{eq:powergrid_eq_theta}-\eqref{eq:powergrid_eq_pf} into stochastic equations.
Then, we use these stochastic equations to estimate the mean and covariances of the power grid states and mechanical wind power injections.
These means and covariances are employed as prior statistics in the GPR equations~\eqref{eq:sef} and \eqref{eq:cef}.

Due to the stochastic nature of wind, we model $P_{k}^m(t)$ as the random process \cite{rosenthal2018ensemble}
\begin{equation}
P_{k}^m(t) = \overline{P}_{k}^m(t) + P_{k}^{\prime m}(t),
\label{windpower_eq}
\end{equation}
with mean $\overline{P}_{k}^m(t) > 0$ and zero-mean Gaussian fluctuations $P_{k}^{\prime m}(t)$ with covariance
\begin{align}
  \label{windcov}
  \left \langle P_{k}^{\prime m}(t) P_{k}^{\prime m}(s) \right \rangle &= \sigma_{k}^2 \exp \left(-\frac{|t-s|}{\lambda_k} \right), \\
  \left \langle P_{k}^{\prime m}(t) P_{l}^{\prime m}(s) \right \rangle &= 0 \quad k \ne l,
\end{align}
where $\sigma_k^2$ and $\lambda_k$ are the variance and correlation time of the fluctuations, respectively.
We assume that the wind prior mean and variance is known (e.g., from meteorological observations).
Following \cite{rosenthal2018ensemble}, we model the fluctuations $P_{k}^{\prime m}(t)$ via the Ornstein-Uhlenbeck (O-U) equation~\cite{arnold1974stochastic}
\begin{equation}
  \label{eq:OU_eq}
  d P_{k}^{\prime m} = -\frac{1}{\lambda_k} P_{k}^{\prime m} d t + \sqrt{\frac{2}{\lambda_k}} \sigma_k dW,
\end{equation}
subject to the initial condition
\begin{equation}
  \label{eq:init_wind}
  P_{k}^{\prime m}(0) = P_{k,0}^{\prime m},
\end{equation}
where $W$ is the standard Wiener process and $P_{k,0}^{\prime m}$ is the random initial condition drawn from the stationary distribution of $P_k^{\prime m}$.

To simulate the system of equations~\eqref{eq:powergrid_eq_theta}-\eqref{eq:powergrid_eq_pf} and \eqref{eq:OU_eq}, we discretize in time using a second-order strong stability preserving Runge-Kutta scheme~\cite{milshtein1994numerical}.
To simplify the notation, we introduce the vectors $y_k = [\theta_k, \omega_k]^T$ and $y = [y^{\top}_1, y^{\top}_2, y^{\top}_3]^{\top}$.
Using this notation, equations~\eqref{eq:powergrid_eq_theta}, \eqref{eq:powergrid_eq_omega}, and~\eqref{eq:OU_eq} read
\begin{align*}
  dy_k &= f_k(y)dt + g_k(y) P_k^{\prime m} dt, \nonumber \\	
  d P_k^{\prime m} &= a_k P_k^{\prime m} d t + b_k dW, \quad k = 1, 2, 3,
\end{align*}
where
\begin{gather*}
  f_k(y)=\begin{bmatrix}
    \omega_B(\omega_k-\omega_s)  \\
    [ \overline{P}_k^m -P_k^e({\bf{\theta }}) - D_k(\omega_k-\omega_s)] / 2 H_k
  \end{bmatrix}, \quad g_k(y)=\begin{bmatrix}
    0  \\
    1 / 2H_k
  \end{bmatrix},\\
  a_k = -\frac{1}{\lambda_k}, \quad b_k = \sqrt{\frac{2}{\lambda_k}} \sigma_k, \quad k = 1, 2, 3.
\end{gather*}

The Runge-Kutta discretization of these equations is~\cite{milshtein1994numerical}
\begin{equation}
  \label{eq:RK_eq}
  \begin{aligned}
    &y_{k,i+1} = y_{k,i} + \frac{h}{2} \{(f_k+g_k P_k^{\prime m} )_i + (f_k+g_k P_k^{\prime m} )_{\bar{i}} \}
    + \frac{1}{\sqrt{12}} g_{k,i} b_k h^{3/2} \eta_i,\\
    &P_{k,i+1}^{\prime m} = P_{k,i}^{\prime m} + b_k \xi_i h^{1/2} + \frac{h}{2} a_k ( P_{k,i}^{\prime m} + P_{k,\bar{i}}^{\prime m} ) + \frac{1}{\sqrt{12}} a_k b_k h^{3/2} \eta_i,
  \end{aligned}
\end{equation}
where $h$ is the time step in the Runge-Kutta scheme satisfying both the stability constraints in~\cite{milshtein1994numerical} and the condition $h = \delta/m$ ($\delta$ is the time between measurements, and $m$ is an integer), $\xi_i$ and $\eta_i$ are the realizations of independent standard Gaussian random variables $\xi$ and $\eta$ at time step $i,$ $f_{k,\bar{i}} = f_k(y_{\bar{i}}),$ $g_{k,\bar{i}} = g_k(y_{\bar{i}}),$ $y_{k,\bar{i}} = y_{k,i} + (f_k + g_k P^{\prime m}_k)_i h$, and $ P_{k,\bar{i}}^{\prime m} = P_{k,i}^{\prime m} + b_k \xi_i h^{1/2} + a_k P_{k,i}^{\prime m} h$.

As described in Section~\ref{sec:phi-gpr-priors}, we use simple Monte Carlo to compute the prior mean and covariances of $\omega_k$,  $\theta_k$, and $P_k^m$ ($k=1,\dots,3$).
Without loss of generality, we assume that $h = \delta$.
Each of the $N_{\mathrm{MC}}$ realizations of the dynamics are generated by sampling~\eqref{eq:RK_eq}.

\section{Simulation results}
\label{sec:results}

Here, we model the three-generator system, shown in Figure~\ref{fig:powergrid}, with generators 1 and 2 powered by wind and generator 3 by traditional power with the known constant mechanical power $P^m_3$. The exact forecast of the power grid with the swing equations is not possible because the mechanical wind power $P^m_1(t)$ and $P^m_2(t)$ are unknown. To compute covariances in the PhI-GPR forecast model, we treat $P^m_1$ and $P^m_2$ as random processes. The three-generator power grid parameters, used in our simulations, are shown in Table \ref{tb1}. 
\begin{table}
\begin{center}
\centering
\caption{System parameters.} \label{tb1}
\begin{tabular}{cccccccc}
   \hline
   $H_1$ $[\mathrm{s}]$ & 13.64 & $H_2$ $[\mathrm{s}]$ & 6.4 &$H_3$ &3.01&$\omega_B$ $[ \mathrm{rad} \mathrm{s}^{-1}]$ &120 \\ \hline
   $D_1$ $[\mathrm{p.u.}]$ & 9.6 &$D_2$ $[\mathrm{p.u.}]$ &2.5&$D_3$ $[\mathrm{p.u.}]$ &1.0&$\omega_s$ $[ \mathrm{rad} \mathrm{s}^{-1}]$ &0 \\ \hline
   $E_1$ $[\mathrm{p.u.}]$ & 1.0156&$E_2$ $[\mathrm{p.u.}]$&1.0359&$E_3$ $[\mathrm{p.u.}]$&1.0053&$\lambda_1$ $[\mathrm{s}]$ &1.8 \\ \hline
   $G_{11}$ $[\mathrm{p.u.}]$ &0.8815 &$G_{21}$ $[\mathrm{p.u.}]$ &0.3083&$G_{31}$ $[\mathrm{p.u.}]$ &0.2258&$\lambda_2$ $[\mathrm{s}]$ &1.8 \\ \hline
   $G_{12}$ $[\mathrm{p.u.}]$ &0.3083 &$G_{22}$ $[\mathrm{p.u.}]$ &0.4357&$G_{32}$ $[\mathrm{p.u.}]$ &0.2247&$\sigma_1$ $[\mathrm{p.u.}]$ &0.05 \\ \hline
   $G_{13}$ $[\mathrm{p.u.}]$ &0.2258 &$G_{23}$ $[\mathrm{p.u.}]$ &0.2247&$G_{33}$ $[\mathrm{p.u.}]$ &0.2860&$\sigma_2$ $[\mathrm{p.u.}]$ &0.05 \\ \hline
   $B_{11}$ $[\mathrm{p.u.}]$ &-3.0273&$B_{21}$ $[\mathrm{p.u.}]$ &1.4904&$B_{31}$ $[\mathrm{p.u.}]$ &1.2088&$\bar{P}_1^m$ $[\mathrm{p.u.}]$ &0.7195 \\ \hline
   $B_{12}$ $[\mathrm{p.u.}]$ &1.4904 &$B_{22}$ $[\mathrm{p.u.}]$ &-2.7397&$B_{32}$ $[\mathrm{p.u.}]$ &1.0764&$\bar{P}_2^m$ $[\mathrm{p.u.}]$ &1.6300 \\ \hline
   $B_{13}$ $[\mathrm{p.u.}]$ &1.2088 &$B_{23}$ $[\mathrm{p.u.}]$ &1.0764&$B_{33}$ $[\mathrm{p.u.}]$ &-2.3770&$\bar{P}_3^m$ $[\mathrm{p.u.}]$ &0.8500 \\
   \hline
\end{tabular}   
\end{center}
\end{table}

Ten thousand Monte Carlo realizations are computed on the time domain $[0,25]$ s with the time step $h=0.0025$ s. The initial conditions are set to $\theta_{1,0}=0.0431$, $\theta_{2,0}=0.4584$, $\theta_{3,0}=0.2372$, and $\omega_{1,0}=\omega_{2,0}=\omega_{3,0}=0$.
Due to the angular indeterminacy of the classical model with constant impedance, we only present phase angles and angular velocities relative to those of the first generator.
\begin{figure}
    \centering
        \includegraphics[scale=.25]{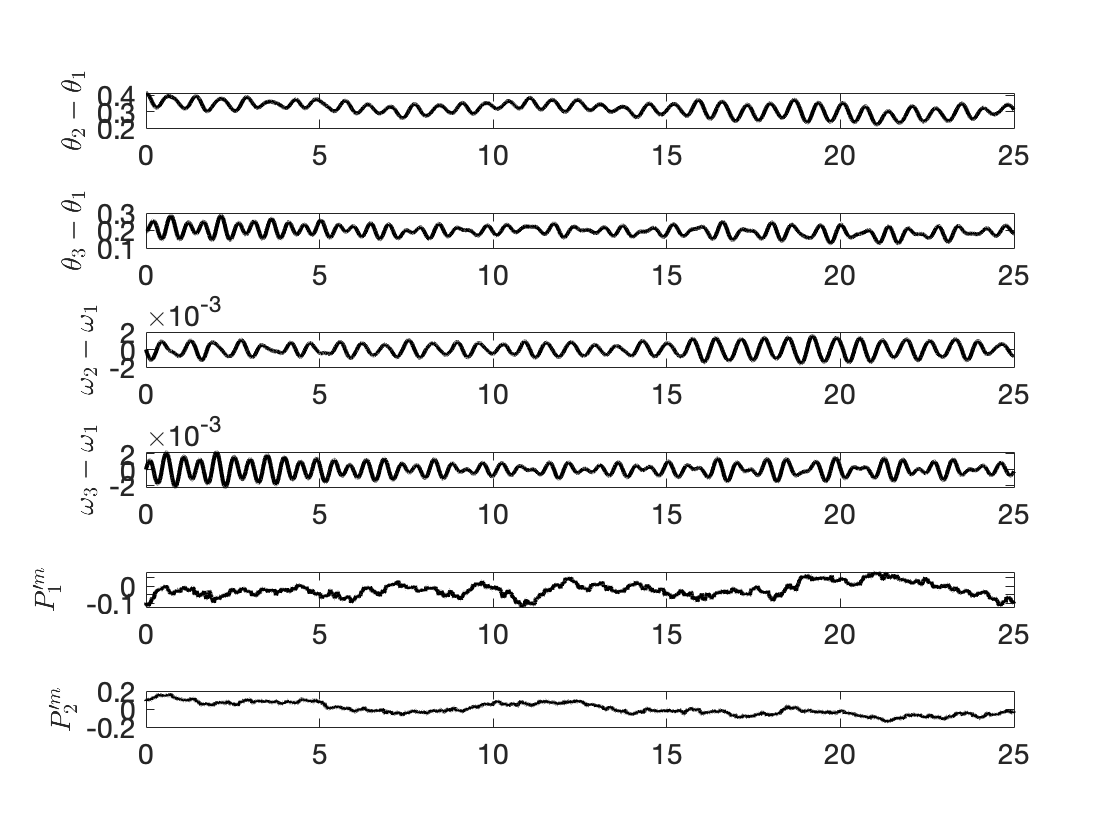}
        \caption{One realization of $\theta_2(t)-\theta_1(t)$, $\theta_3(t)-\theta_1(t)$, $\omega_2(t)-\omega_1(t)$, $\omega_3(t)-\omega_1(t)$, $P_1^{\prime m}(t)$, and $P_2^{\prime m}(t)$.} \label{onerealization_states}
\end{figure} 
\begin{figure}
    \centering
        \includegraphics[scale=.25]{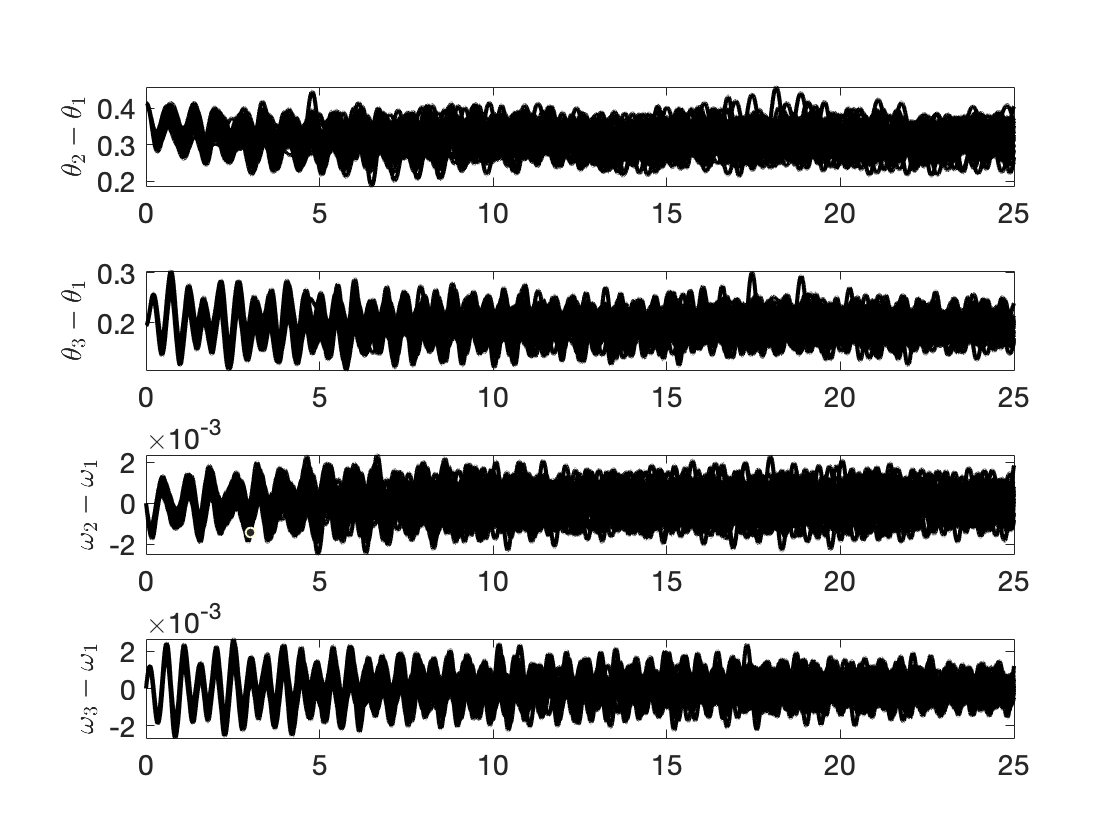}
        \caption{Fifty realizations of $\theta_2(t)-\theta_1(t)$, $\theta_3(t)-\theta_1(t)$, $\omega_2(t)-\omega_1(t)$, and $\omega_3(t)-\omega_1(t)$.} \label{50realizations_states}
\end{figure}    
\begin{figure}
    \centering
        \includegraphics[scale=.25]{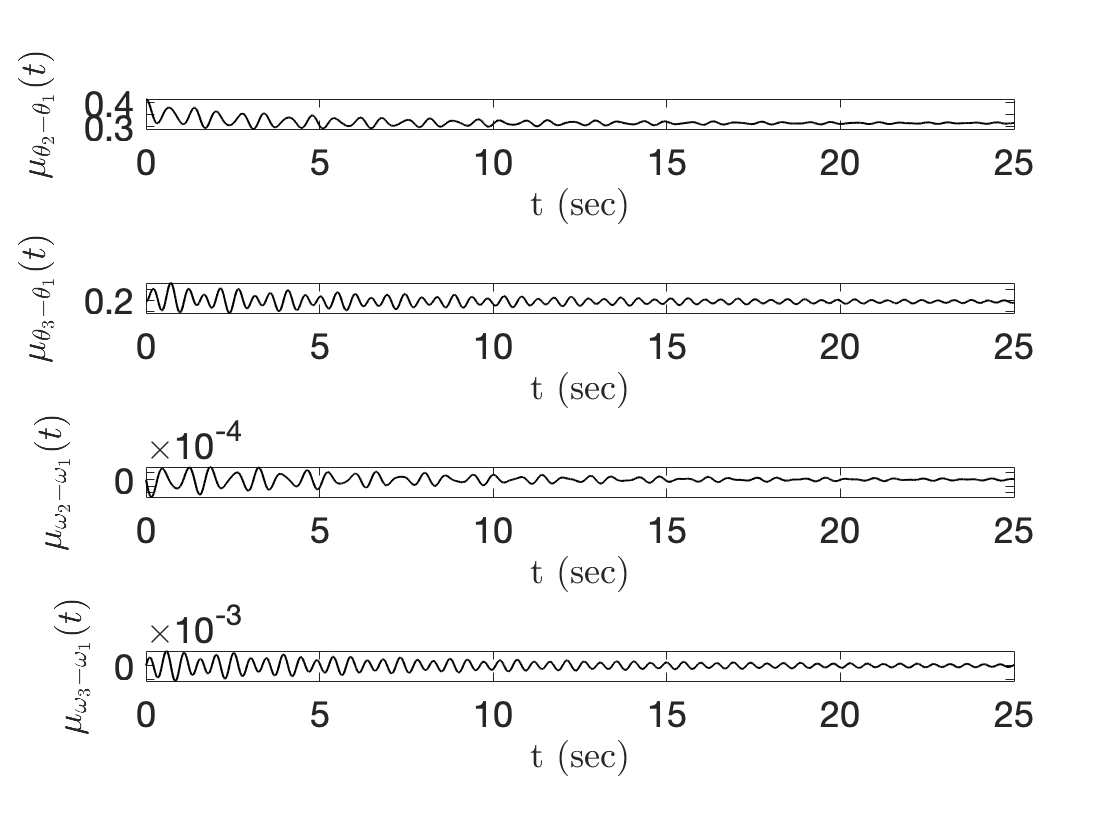}
        \caption{Mean of $\theta_2(t)-\theta_1(t)$, $\theta_3(t)-\theta_1(t)$, $\omega_2(t)-\omega_1(t)$, and $\omega_3(t)-\omega_1(t)$.} \label{mean_states}
\end{figure}
\begin{figure}
    \centering
        \includegraphics[scale=.25]{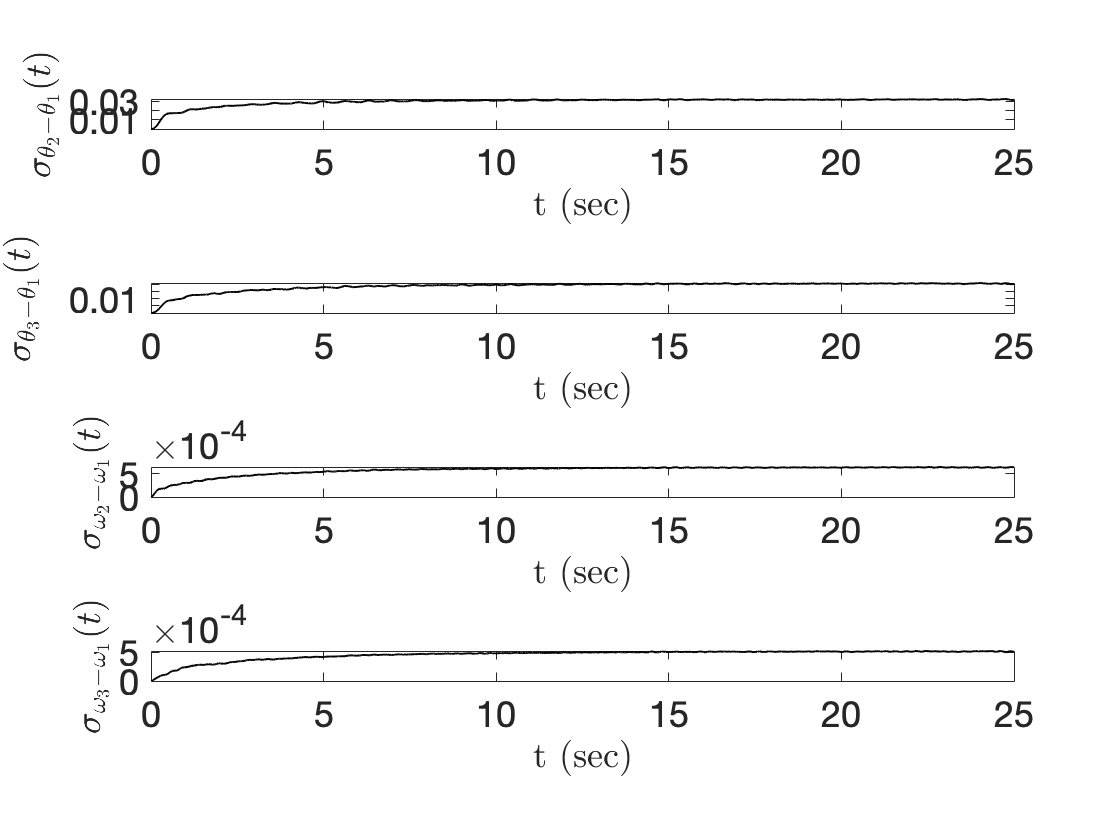}
        \caption{Standard deviation of $\theta_2(t)-\theta_1(t)$, $\theta_3(t)-\theta_1(t)$, $\omega_2(t)-\omega_1(t)$, and $\omega_3(t)-\omega_1(t)$.}\label{cov_states}
\end{figure}

Figure \ref{onerealization_states} shows one realization of $\theta_2(t)-\theta_1(t)$, $\theta_3(t)-\theta_1(t)$, $\omega_2(t)-\omega_1(t)$, $\omega_3(t)-\omega_1(t)$, $P_1^{\prime m}(t)$, and $P_2^{\prime m}(t)$ that we randomly select from the ensemble and use as the ground truth in the numerical examples below. 

In Figure \ref{50realizations_states},  50  realizations of $\theta_2(t)-\theta_1(t)$, $\theta_3(t)-\theta_1(t)$, $\omega_2(t)-\omega_1(t)$, and $\omega_3(t)-\omega_1(t)$ are presented to demonstrate variability in the states caused by the  randomly varying $P^m_k(t)$. The mean and standard deviation of $\theta_2(t)-\theta_1(t)$, $\theta_3(t)-\theta_1(t)$, $\omega_2(t)-\omega_1(t)$, and $\omega_3(t)-\omega_1(t)$  are shown in figures \ref{mean_states} and \ref{cov_states}, which demonstrate that the states converge to a statistical steady state (constant mean and standard deviation of states)  after approximately 15 s. In this work, we focus on the forecast of states for times less than 15.5 s, i.e., the forecast of states with non-stationary (evolving) statistics. Such predictions are especially challenging for the standard GPR, which relies on the assumption of stationary statistics for the covariance estimation. Specifically, we set $T_0$ to $8.3375$ s and $T_f$ to $12.5$ s. We note that to obtain the PhI-GPR forecast until time $T_f=12.5$ s, we only need to obtain Monte Carlo solution for the covariance on the time interval $[0,T_f]$. Here, we obtain a solution on a larger time domain only to demonstrate the nonstationarity of prior statistics.   

In Section \ref{s1}, we assume that measurements are only available for $\theta_k$ ($k=1 ,2, 3$), and in Section \ref{sec:omega}, we assume that  only $\omega_k$ ($k=1 ,2, 3$) measurements are available. In both cases, we forecast all states $\theta_2(t)-\theta_1(t)$, $\theta_3(t)-\theta_1(t)$, $\omega_2(t)-\omega_1(t)$, $\omega_3(t)-\omega_1(t)$, $P_1^{\prime m}(t)$ and $P_2^{\prime m}(t)$ and estimate the unobserved states. In Section \ref{data-driven GPR}, we compare the PhI-GPR and the standard data-driven GPR when measurements of both $\theta_k$ and $\omega_k$ ($k=1 ,2, 3$) are available. In Section \ref{noiseless} and \ref{noisy}, we compare the PhI-GPR and ARIMA forecasts of $\theta_k$ and $\omega_k$ using both noiseless and noisy measurements of  $\theta_k$ and $\omega_k$. 

\subsection{PhI-GPR state estimation and forecasting using measurements of $\theta_k$ ($k=1 ,2, 3$)}\label{s1}
\begin{figure}
    \centering
        \includegraphics[scale=.25]{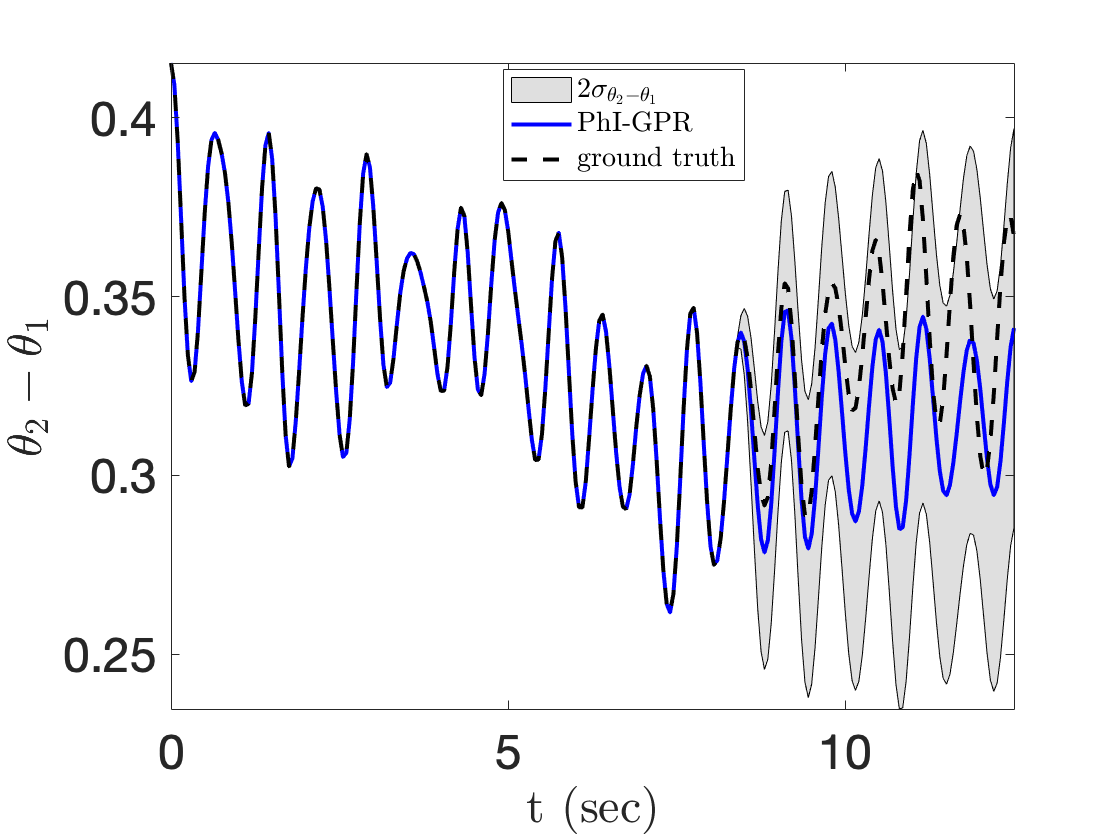}
        \caption{Forecasting of $\theta_2(t)-\theta_1(t)$ when measurements of $\theta_k$ ($k=1 ,2, 3$) are available for $t<8.3375$ s every $0.05 $ s.}\label{tt_PhI_GPR}
\end{figure}
\begin{figure}
    \centering
        \includegraphics[scale=.25]{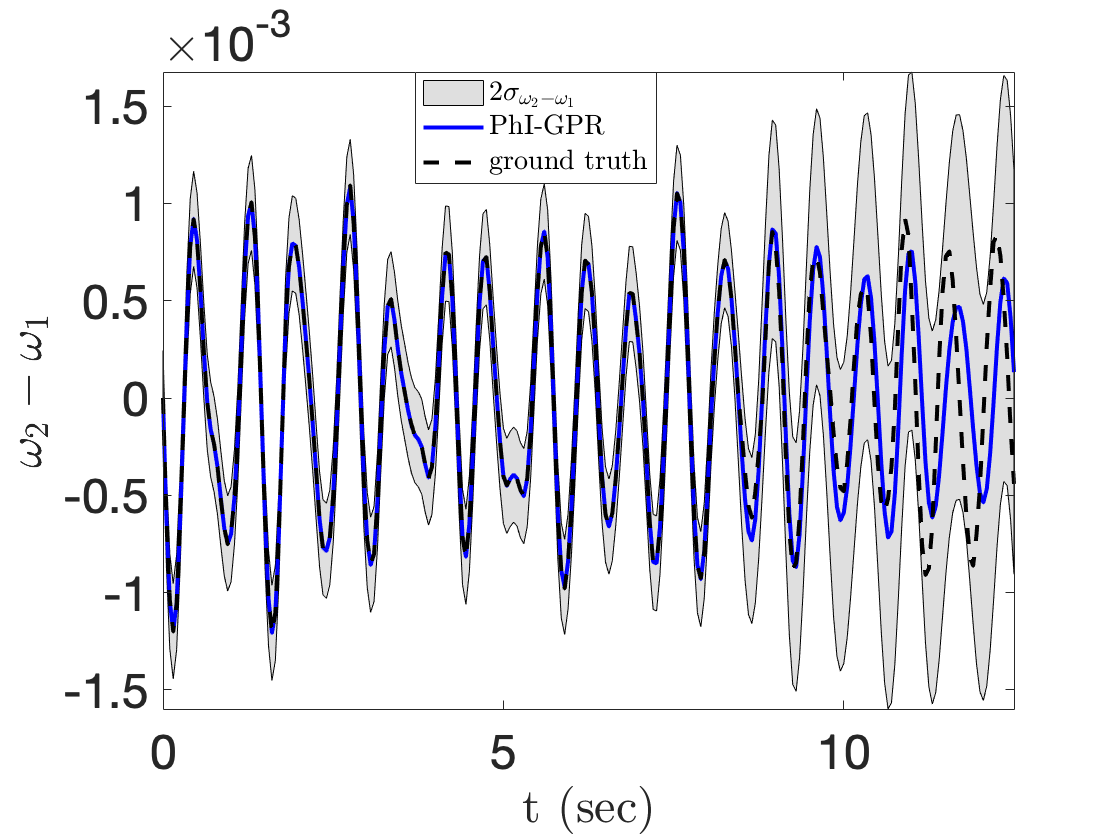}
        \caption{Forecasting of  $\omega_2(t)-\omega_1(t)$ when measurements of $\theta_k$ ($k=1 ,2, 3$) are available for $t<8.3375$ s every $0.05$ s.}\label{to_PhI_GPR}
\end{figure}
\begin{figure}
    \centering
        \includegraphics[scale=.25]{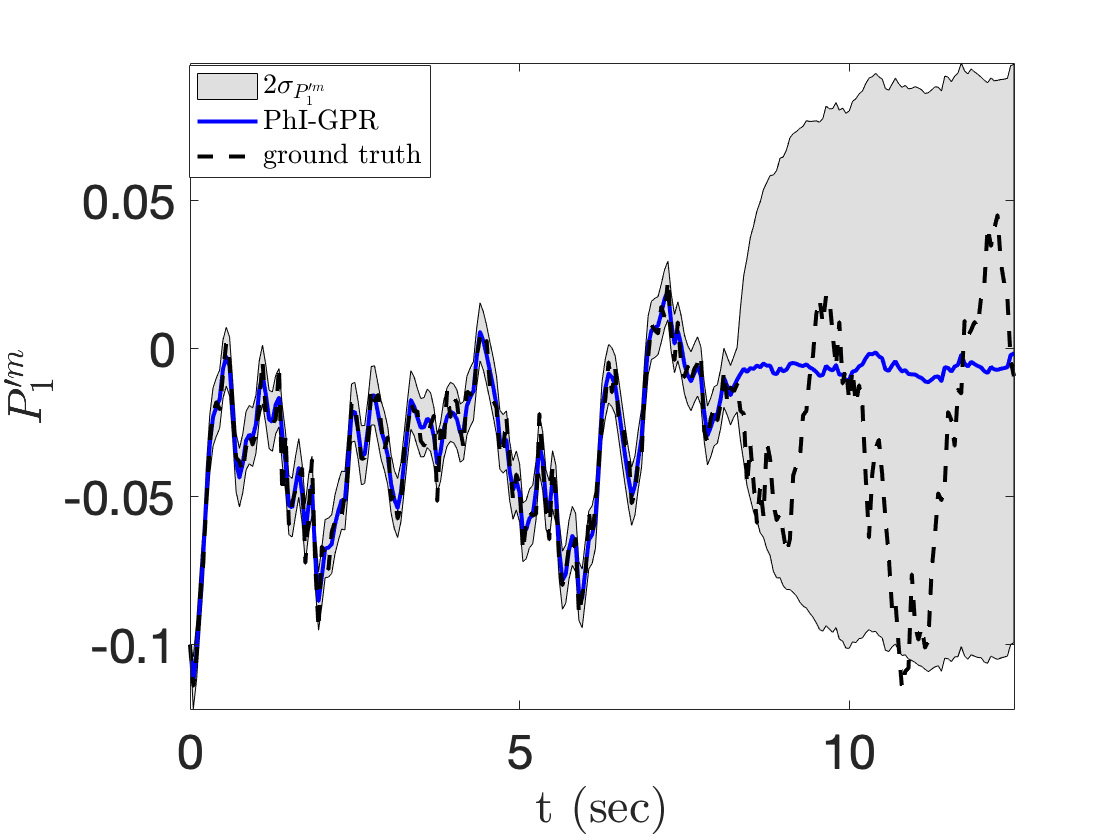}
        \caption{Forecasting of $P_1^{\prime m}(t)$ when measurements of $\theta_k$ ($k=1 ,2, 3$) are available for $t<8.3375$ s every $0.05$ s.} \label{tp_PhI_GPR}
\end{figure} 
In this case, we assume that measurements of $\theta_k$ ($k=1 ,2, 3$) are available for $t<8.3375$ s every $0.05$ s. Our goal is to predict $\theta_2(t)-\theta_1(t)$ and $\theta_3(t)-\theta_1(t)$ for $t>8.3375$ s and $\omega_2(t)-\omega_1(t)$, $\omega_3(t)-\omega_1(t)$, $P_1^{\prime m}(t)$, and $P_2^{\prime m}(t)$ for the entire time interval $t\in[0,12.5]$ s. The measurements are taken from the ground truth solution for $t<8.3375$ s. 
The ground truth solution is also used to validate the PhI-GPR forecast of $\theta_2(t)-\theta_1(t)$ and $\theta_3(t)-\theta_1(t)$ for $t>8.3375$ s and  $\omega_2(t)-\omega_1(t)$, $\omega_3(t)-\omega_1(t)$,  $P_1^{\prime m}(t)$, and $P_2^{\prime m}(t)$ for $t\in[0,12.5]$ s.
In the PhI-GPR forecast, the prior mean and covariance of states are computed as described in Section~\ref{sec:phi-gpr-priors} using the stochastic model presented in Section~\ref{sec:phi-gpr-power-grid}.
We draw $10^4$ realizations of the stochastic model.
One of these realizations is used as the ground truth, while the remaining $10^4 - 1$ realizations are employed to compute the prior mean and covariances for PhI-GPR via simple Monte Carlo.

In Table \ref{tb2}, we give the log predictive probabilities for the states $\theta_2(t)-\theta_1(t)$, $\theta_3(t)-\theta_1(t)$, $\omega_2(t)-\omega_1(t)$, $\omega_3(t)-\omega_1(t)$, $P_1^{\prime m}(t)$, and $P_2^{\prime m}(t)$.
The log predictive probability is a quantitative measure of the accuracy of predictions from statistical models, which corresponds to the sum of the pointwise log probabilities of reference values being observed given the statistical model~\cite{williams2006gaussian}.
For a certain estimated or forecasted state $\alpha(t)$, it is given by
\begin{equation*}
  \text{log predictive probability} = - \sum^{N^{f/e}}_{k = 1} \left \{ \frac{ \left [ \mu^{f/e}(t_k) - \alpha(t_k) \right ]^2}{2 \left [ \sigma^{f/e}(t_k) \right ]^2} + \frac{1}{2} \log 2 \pi \left [ \sigma^{f/e}(t_k) \right ]^2 \right \}
\end{equation*}
where $N^{f/e}$ denotes the number of forecast or estimation times, $\mu^{f/e}(t_k)$ and $\sigma^{f/e}(t_k)$ are the posterior mean and standard deviation of the forecast or estimation at time $t_k$, respectively, and $\alpha(t_k)$ is the reference value at time $t_k$.
The larger the log predictive probability, the more accurate is the model estimation or forecast.

\begin{table}
\begin{center}
\centering
\caption{Log predictive probabilities for $\theta_2(t)-\theta_1(t)$, $\theta_3(t)-\theta_1(t)$, $\omega_2(t)-\omega_1(t)$, $\omega_3(t)-\omega_1(t)$, $P_1^{\prime m}(t)$, and $P_2^{\prime m}(t)$ when measurements of $\theta_k$ ($k=1,2,3$) are available for $t<T_0=8.3375$ s every $0.05$ s.} \label{tb2}

\begin{tabular}{ |c |c | c |c |c | c |c| }
   \hline
   \multicolumn{1}{|c}{\multirow{2}{*}{Generator $k$}} & 
   \multicolumn{2}{|c|}{$\theta_k-\theta_1$} & 
   \multicolumn{2}{c|}{$\omega_k-\omega_1$}&
   \multicolumn{2}{c|}{$P_k^{\prime m}$}  \\ \cline{2-7}
   \multicolumn{1}{|c}{} &
   \multicolumn{1}{|c|}{$t<T_0$}& $t>T_0$ & $t<T_0$ & $t>T_0$ & $t<T_0$ & $t>T_0$ \\ \hline
   1 & NA & NA & NA & NA & 663.302 & 136.435 \\ \hline
   2 & NA & 194.642 & 1359.04 & 550.43 & 690.176 & 124.035 \\ \hline
   3 & NA & 247.571 & 1359.29 & 573.736 & NA & NA \\  \hline
\end{tabular}  
\end{center} 
\end{table}

Figures \ref{tt_PhI_GPR} -- \ref{tp_PhI_GPR} show the state estimations and forecasts and the associated uncertainties.  The forecast of $\theta_2(t)-\theta_1(t)$ and $\omega_2(t)-\omega_1(t)$  for $t\in[8.3,12.5]$ s is satisfactory, with the ground truth staying within two standard deviations of the predicted states. For the first two seconds (approximately the correlation time of the states), the forecasted value matches the ground truth very closely. The forecast of $P^{\prime m}_1$ is less satisfactory, but the ground truth still mostly stays within two standard deviations of the forecasted value. The challenges with forecasting $P^{\prime m}_k$ are to be expected due to its stochastic nature. The estimation of the unobserved states ($\omega_2(t)-\omega_1(t)$ and $P^{\prime m}_1$ for $t<8.3$ s   is very accurate, indicating that there is a very strong correlation between the observed and unobserved states. Another outcome of the strong correlation is that the uncertainty (the predicted standard deviation) for parameter estimation (for $t<8.3$ s) is much smaller than that for forecasting (for $t>8.3$ s). 
The forecast and estimation of the remaining states show similar behavior and, for this reason, are not presented here. 

\subsection{PhI-GPR forecasting using measurements of $\omega_k$ ($k=1 ,2, 3$)} \label{sec:omega}
\begin{figure}
    \centering
        \includegraphics[scale=.25]{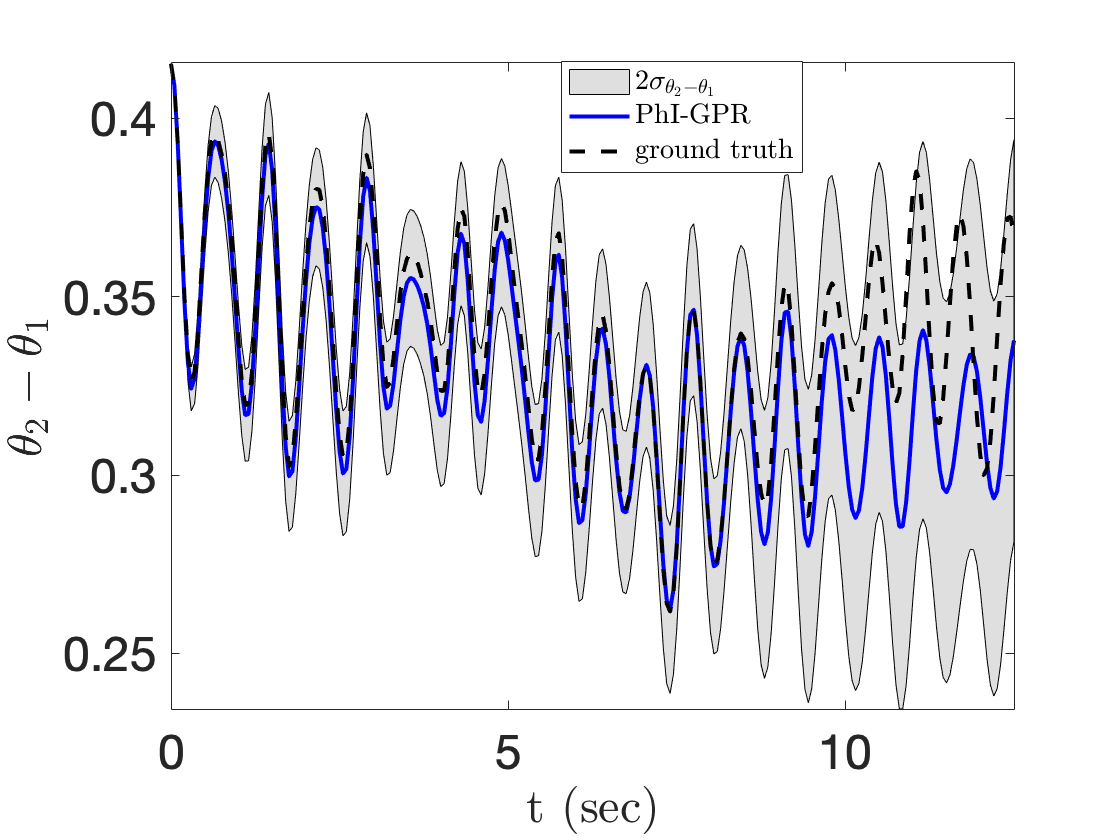}
        \caption{Forecasting of $\theta_2(t)-\theta_1(t)$ when measurements of $\omega_k$ ($k=1 ,2, 3$) are available for $t<8.3375$ s every $0.05 $ s.}\label{ot_PhI_GPR}
\end{figure}
\begin{figure}
    \centering
        \includegraphics[scale=.25]{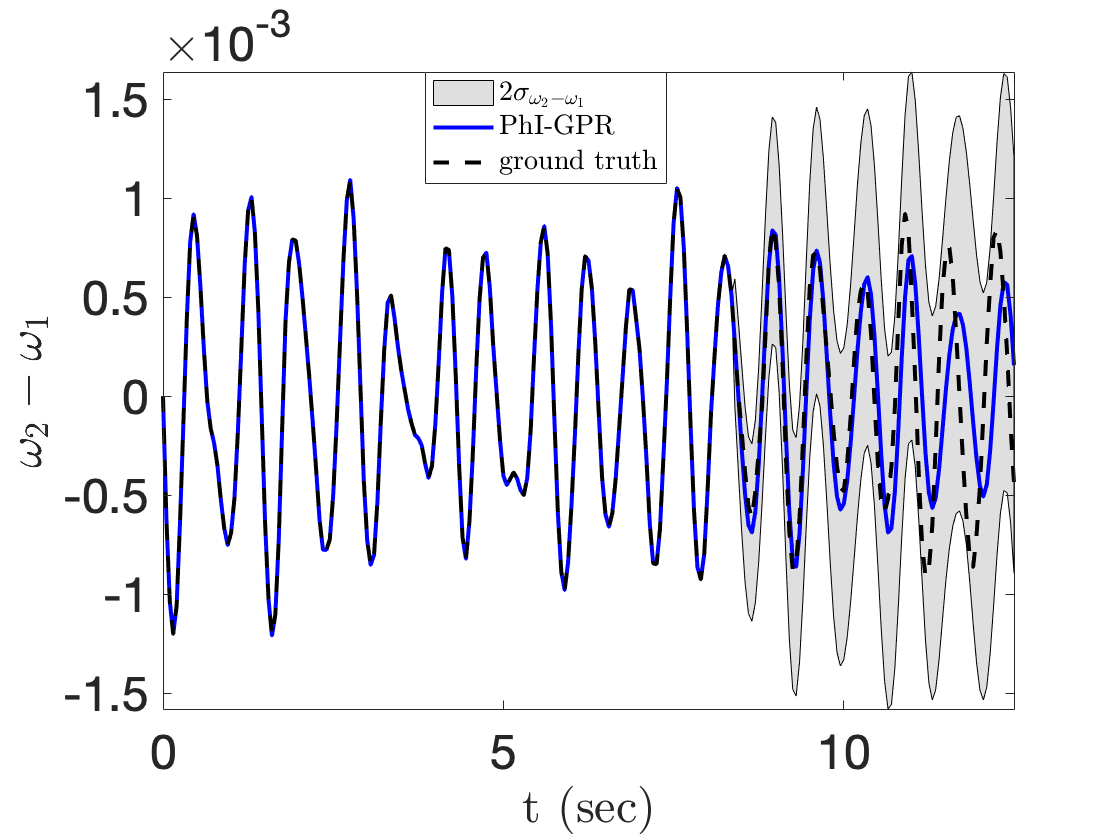}
        \caption{Forecasting of  $\omega_2(t)-\omega_1(t)$ when measurements of $\omega_k$ ($k=1 ,2, 3$) are available for $t<8.3375$ s every $0.05$ s.}\label{oo_PhI_GPR}
\end{figure}
\begin{figure}
    \centering
        \includegraphics[scale=.25]{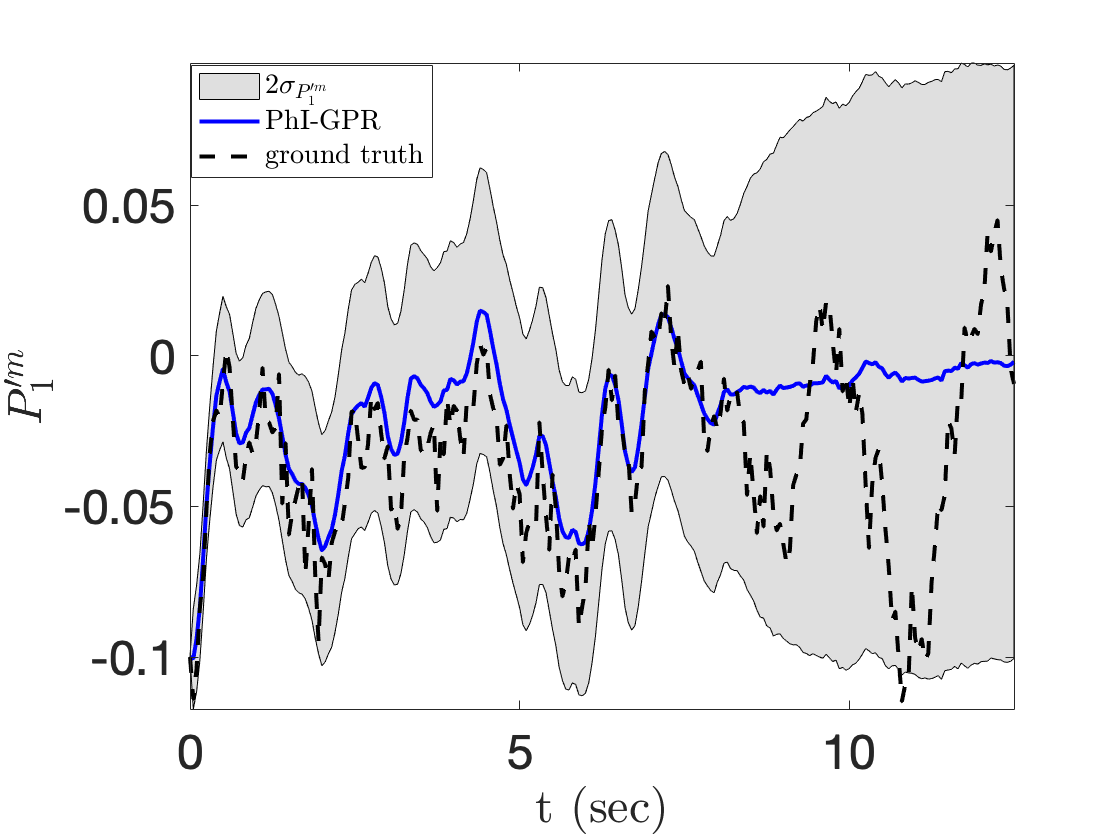}
        \caption{Forecasting $P_1^{\prime m}(t)$ when measurements of $\omega_k$ ($k=1 ,2, 3$) are available for $t<8.3375$ s every $0.05$ s.} \label{op_PhI_GPR}
\end{figure} 

Here, we assume that measurements of $\omega_k$ ($k=1 ,2, 3$) are available for $t<8.3375$ s every $0.05$ s. Our goal is to predict $\omega_2(t)-\omega_1(t)$ and $\omega_3(t)-\omega_1(t)$ for $t>8.3375$ s and $\theta_2(t)-\theta_1(t)$, $\theta_3(t)-\theta_1(t)$, $P^{\prime m}_{1}(t)$, and $P^{\prime m}_{2}(t)$ for the entire time interval $t\in[0,12.5] $ s. As before, the ground truth solution provides measurements of observed states and is used to validate forecasted and estimated states. The log predictive probabilities for $\theta_2(t)-\theta_1(t)$, $\theta_3(t)-\theta_1(t)$, $\omega_2(t)-\omega_1(t)$, $\omega_3(t)-\omega_1(t)$, $P_1^{\prime m}(t)$, and $P_2^{\prime m}(t)$ are given in Table \ref{tb3}. 

\begin{table}
\begin{center}
\centering
\caption{Log predictive probabilities for $\theta_2(t)-\theta_1(t)$, $\theta_3(t)-\theta_1(t)$, $\omega_2(t)-\omega_1(t)$, $\omega_3(t)-\omega_1(t)$, $P_1^{\prime m}(t)$, and $P_2^{\prime m}(t)$ when measurements of $\omega_k$ ($k=1 ,2, 3$) are available for $t<T_o=8.3375$ s every $0.05$ s.} \label{tb3}

\begin{tabular}{ |c |c | c |c |c | c |c| }
   \hline
   \multicolumn{1}{|c}{\multirow{2}{*}{Generator $k$}} & 
   \multicolumn{2}{|c|}{$\theta_k-\theta_1$} & 
   \multicolumn{2}{c|}{$\omega_k-\omega_1$}&
   \multicolumn{2}{c|}{$P_k^{\prime m}$}  \\ \cline{2-7}
   \multicolumn{1}{|c}{} &
   \multicolumn{1}{|c|}{$t<T_0$}& $t>T_0$ & $t<T_0$ & $t>T_0$ & $t<T_0$ & $t>T_0$ \\ \hline
   1 & NA & NA & NA & NA & 463.01 & 140.327 \\ \hline
   2 & 624.416 & 183.523 & NA & 549.362 & 489.118 & 115.266 \\ \hline
   3 & 735.93 & 240.512 & NA & 570.322 & NA & NA \\  \hline
\end{tabular}  
\end{center} 
\end{table}

Figure \ref{ot_PhI_GPR} shows the prediction of the non-observed states $\theta_2 - \theta_1$. The estimation of $\theta_2 - \theta_1$ for  $t<8.3375$ s is very accurate. The accurate forecast (where the GPR prediction closely tracks the ground truth) is at least 2 s. Later, the accuracy decreases but the ground truth remains within two times the standard deviations of the predicted values. We see the similarly accurate forecast of the observed states $\omega_2 - \omega_1$ in Figure \ref{oo_PhI_GPR}.    

Figure \ref{op_PhI_GPR} demonstrates that the estimation of $P^{\prime m}_{1}(t)$ for  $t<8.3375$ s based on $\omega_k$ observations is not as good as that based on $\theta_k$ measurements in Section \ref{s1}, indicating that $P^{\prime m}_{1}(t)$ is more strongly correlated to $\theta_k$ than $\omega_k$.  The accuracy of the $P^{\prime m}_{1}(t)$  forecast for $t>8.3375$ s based on $\omega_k$ measurements is approximately the same as that based on $\theta_k$ measurements, with the ground truth being within two standard deviations of the PhI-GPR predicted values. 

\subsection{Comparison between physics-informed and data-driven GPR methods}\label{data-driven GPR}
\begin{figure}
    \centering
        \includegraphics[scale=.25]{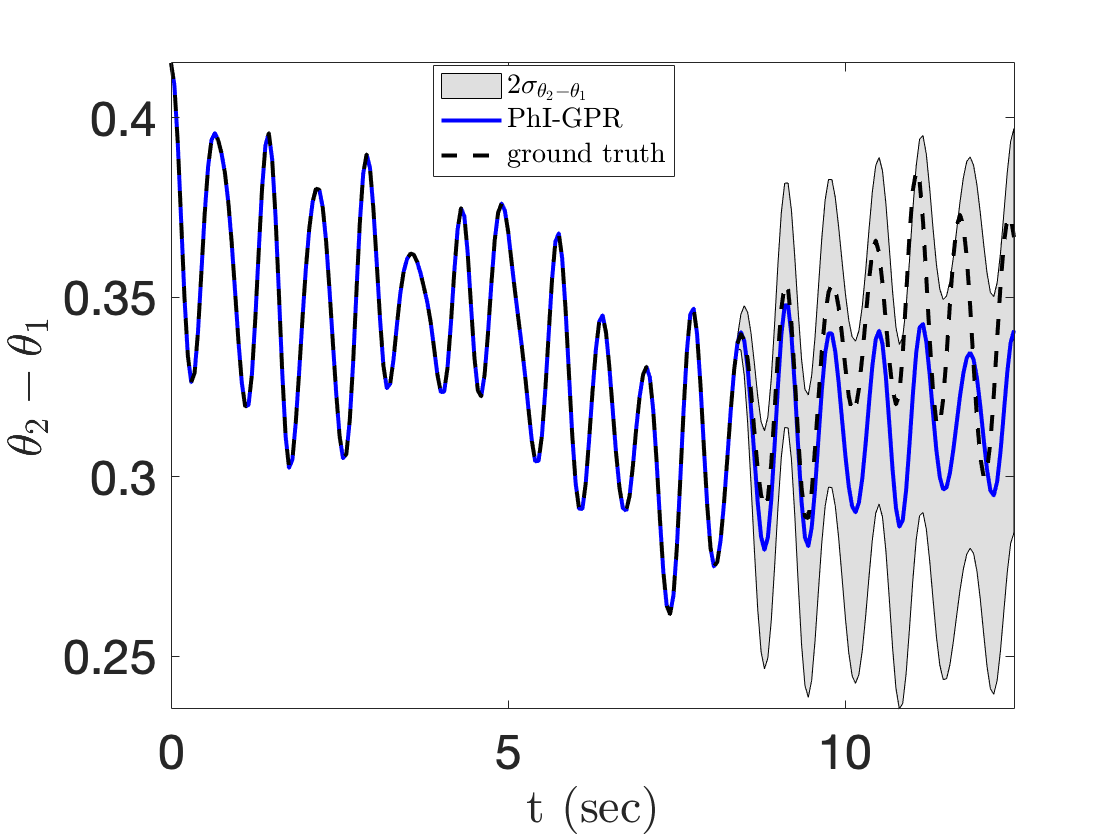}
        \caption{PhI-GPR forecast of $\theta_2(t)-\theta_1(t)$ with $\theta_k$ and $\omega_k$ ($k=1 ,2, 3$) measurements available for $t<8.3375$ s every $0.05$ s.}\label{tot_PhI_GPR}
\end{figure}
\begin{figure}
    \centering
        \includegraphics[scale=.25]{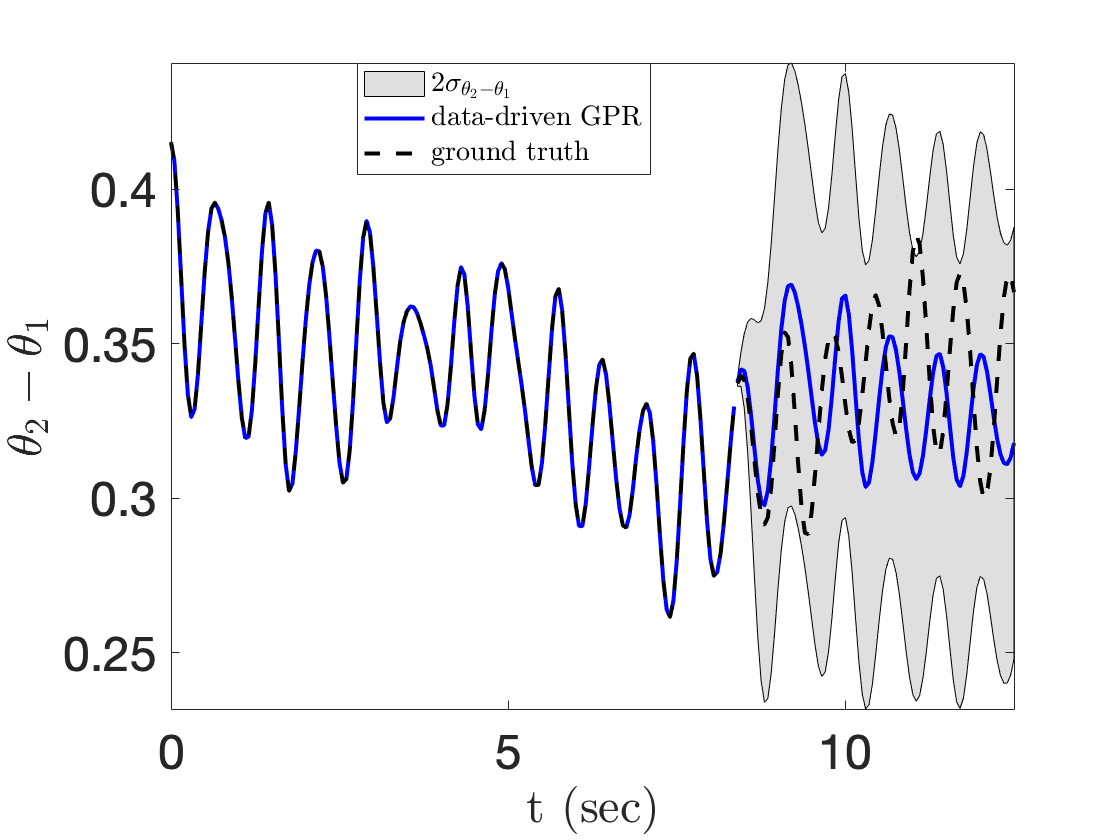}
        \caption{Data-driven GPR forecast of $\theta_2(t)-\theta_1(t)$ with measurements of $\theta_k$ and $\omega_k$ ($k=1 ,2, 3$) for $t<8.3375$ s available every $0.05$ s.}\label{tot_GPR}
\end{figure}

In this section, we provide comparisons between PhI-GPR and the standard data-driven GPR for a case where the measurements of both $\theta_k$ and $\omega_k$ ($k=1 ,2, 3$) are available for $t<8.3375$ s every $0.05$ s. Our goal is to predict $\theta_2(t)-\theta_1(t)$, $\theta_3(t)-\theta_1(t)$, $\omega_2(t)-\omega_1(t)$, and $\omega_3(t)-\omega_1(t)$ for $t>8.3375 $ s.

In the data-driven GPR, we assume the covariance function models given by a combination of the squared exponential function, rational quadratic function, periodic function, and Kronecker delta function as:
\begin{multline*}
  K_{\alpha \alpha}(t, \tau) = \gamma^2_1 \exp \left [ - \frac{(t - \tau)^2}{2 \gamma^2_2} \right ] + \gamma^2_3 \left [ 1 + \frac{(t - \tau)^2}{2 \gamma_4 \gamma^2_5} \right ]^{-\gamma_4}\\
  + \gamma^2_6 \exp \left \{ - \frac{2 \sin^2 \left [ \frac{\pi}{24} (t - \tau) \right ]}{\gamma^2_7} \right \} + \gamma^2_8 \delta(t - \tau).
\end{multline*}
The means $\overline{\omega}^o(t) = \overline{\omega}^f(t)$ and $\overline{\theta}^o(t)=\overline{\theta}^f(t)$ and parameters $\gamma_i$ ($i = 1, \dots, 8$) are determined by minimizing the negative marginal likelihood of the observed data~\cite{williams2006gaussian}.

The log predictive probabilities for the forecasted values of $\theta_2(t)-\theta_1(t)$, $\theta_3(t)-\theta_1(t)$, $\omega_2(t)-\omega_1(t)$, and $\omega_3(t)-\omega_1(t)$, obtained with the PhI-GPR and data-driven GPR, are given in tables \ref{tb4} and \ref{tb5}, respectively.  

\begin{table}
\begin{center}
\centering
\caption{Log predictive probabilities for $\theta_2(t)-\theta_1(t)$, $\theta_3(t)-\theta_1(t)$, $\omega_2(t)-\omega_1(t)$, and $\omega_3(t)-\omega_1(t)$ using PhI-GPR when measurements of $\theta_k$ and $\omega_k$ ($k=1 ,2, 3$) are available for $t<T_0 = 8.3375$ s every $0.05$ s. } \label{tb4}

\begin{tabular}{ |c |c | c |c |c |}
   \hline
   \multicolumn{1}{|c}{\multirow{2}{*}{Generator $k$}} & 
   \multicolumn{2}{|c|}{$\theta_k-\theta_1$} & 
   \multicolumn{2}{c|}{$\omega_k-\omega_1$} \\ \cline{2-5}
   \multicolumn{1}{|c}{} &
   \multicolumn{1}{|c|}{$t<T_0$}& $t>T_0$ & $t<T_0$ & $t>T_0$  \\ \hline
   1 & NA & NA & NA & NA  \\ \hline
   2 &NA & 198.858 & NA & 551.258 \\ \hline
   3 & NA & 246.031 & NA & 565.729 \\  \hline
\end{tabular}  
\end{center} 
\end{table}

\begin{table}
\begin{center}
\centering
\caption{Log predictive probabilities for $\theta_2(t)-\theta_1(t)$, $\theta_3(t)-\theta_1(t)$, $\omega_2(t)-\omega_1(t)$, and $\omega_3(t)-\omega_1(t)$ using data-driven GPR when measurements of $\theta_k$ and $\omega_k$ ($k=1 ,2, 3$) are available for $t<T_0 = 8.3375$ s every $0.05$ s. } \label{tb5}

\begin{tabular}{ |c |c | c |c |c |}
   \hline
   \multicolumn{1}{|c}{\multirow{2}{*}{Generator $k$}} & 
   \multicolumn{2}{|c|}{$\theta_k-\theta_1$} & 
   \multicolumn{2}{c|}{$\omega_k-\omega_1$} \\ \cline{2-5}
   \multicolumn{1}{|c}{} &
   \multicolumn{1}{|c|}{$t<T_0$}& $t>T_0$ & $t<T_0$ & $t>T_0$  \\ \hline
   1 & NA & NA & NA & NA  \\ \hline
   2 &NA & 211.542 & NA & 555.884 \\ \hline
   3 & NA & 261.263 & NA & 515.667 \\  \hline
\end{tabular}  
\end{center} 
\end{table}

Figures \ref{tot_PhI_GPR} and \ref{tot_GPR} show the forecast of $\theta_2(t)-\theta_1(t)$ using the PhI-GPR and standard data-driven GPR, respectively.   
For these states, the log predictive probabilities of the data-driven GPR forecast are slightly larger (less than 10\%) than those of the PhI-GPR forecast. However, figures \ref{tot_PhI_GPR} and \ref{tot_GPR}  clearly show that the PhI-GPR forecast is significantly more accurate than the data-driven GPR forecast for the first 2 s. 
The PhI-GPR forecast closely matches the ground truth for approximately 2 s. After that, the ground truth stays within two  standard deviations of the GPR prediction. 
The data-driven GPR forecast deviates from the exact forecast after approximately 1 s, but also stays within two standard deviations of the ground truth. We note that the data-driven GPR results in a less certain forecast, i.e., the data-driven GPR produces a larger standard deviation of forecasted variables than the PhI-GPR.

\subsection{Comparison between PhI-GPR and  ARIMA methods}\label{noiseless}
In this section, we compare PhI-GPR with the univariate ARIMA method. Unlike PhI-GPR, the univariate ARIMA method only allows the forecasting of observed variables, e.g., $\theta_k$ measurements are needed for  forecasting $\theta_k$. As in GPR, the ARIMA forecast of $\theta_k^f$ and $\omega_k^f$ is given as a linear combination of the $N_o$ measurements of $\theta_k^o$ and $\omega_k^o$, respectively.
Using the notation introduced in Section~\ref{sec:multivariate-GPR}, the ARIMA forecast can be expressed as 
\begin{align}
\omega _k^f&(t_{N_o+1}) - \alpha _{k,N_o} \omega _k^o (t_{N_o}) -  \alpha _{k,N_o-1} \omega _k^o (t_{N_o-1})-  \cdots  -  \alpha _{k,p_1} \omega _k^o (t_{N_o-p_1})  \\ \nonumber
&= {e_{k,N_o+1}} + \beta _{k,N_o}  e_{k,N_o} + \beta _{k,N_o-1}  e_{k,N_o-1} +   \cdots  + \beta_{k,N_o-q _1}  e_{k,N_o-q _1} \\
\theta _k^f&(t_{N_o+1}) - \gamma _{k,N_o} \theta_k^o (t_{N_o}) -  \gamma_{k,N_o-1} \theta_k^o (t_{N_o-1})-  \cdots  -  \gamma_{k,p_2} \theta_k^o (t_{N_o - p_2})  \\ \nonumber
&= {\epsilon_{k,N_o+1}} + {\lambda_{k, N_o}}{\epsilon_{k, N_o}} + {\lambda_{k, N_o-1}}{\epsilon_{k, N_o-1}} +  \cdots  + \lambda _{k, N_o-q_2}  \epsilon_{k, N_o -q_2}
\end{align} 
where $\omega _k^f(t_{N_o+1})$ and $\theta _k^f(t_{N_o+1})$ are forecasted values of $\omega$ and $\theta$, respectively. 
The coefficients 
$p_1$, $p_2$ and $\alpha_{k,i}$ and $\gamma_{k,i}$ are the orders and parameters of the autoregressive part, respectively; $q_1$, $q_2$ and $\beta_{k,i}$ and $\lambda_{k,i}$ are the orders and parameters of the moving averaging part, respectively; and $e_{k,i}$ and $\epsilon_{k,i}$ are zero mean independent normally distributed error terms. Non-seasonal ARIMA models are generally denoted as ARIMA$(p,d,q)$, where $d$ is the degree of differencing used to remove a trend in data. Here, we assume that the data does not have a trend and set $d$ to zero. The order parameters are selected using the Akaike's Information Criterion~\cite{hyndman2018forecasting}.
 
\begin{figure}
    \centering
        \includegraphics[scale=.25]{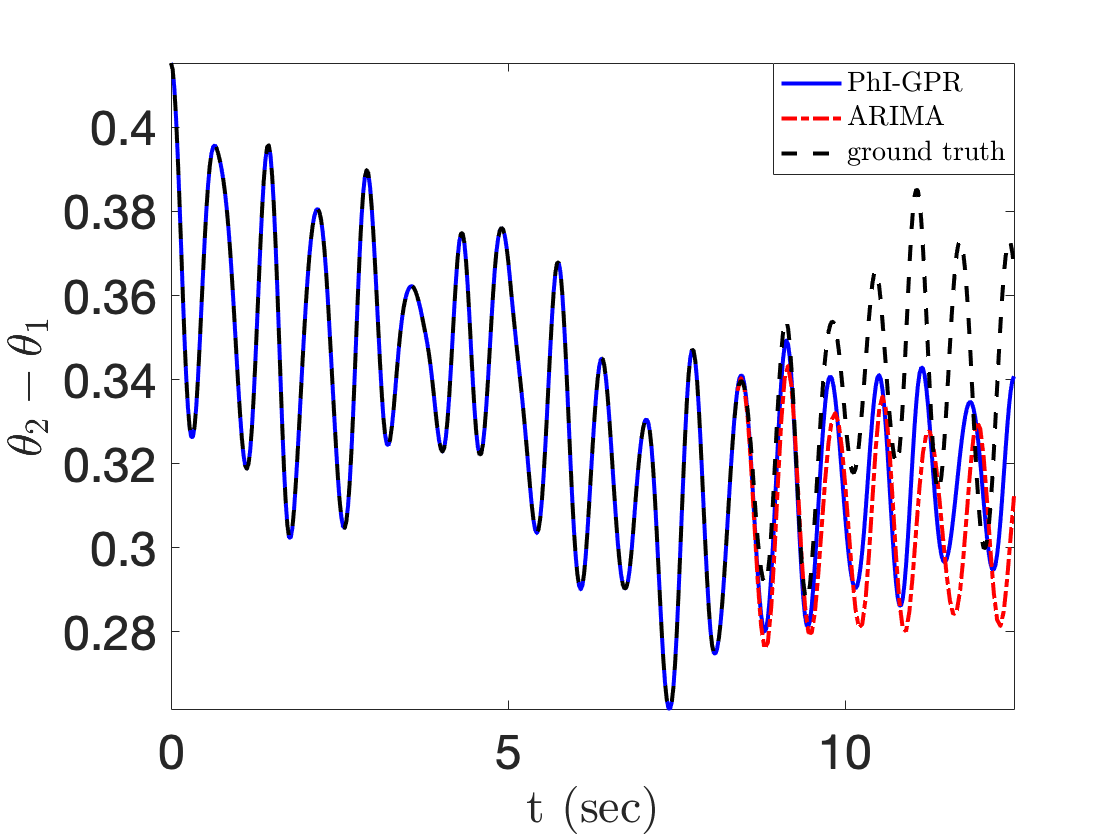}
        \caption{Forecasting of $\theta_2(t)-\theta_1(t)$ using GPR and ARIMA when measurements of $\theta_k$ and $\omega_k$ ($k=1 ,2, 3$) are available for $t<8.3375$ s every $0.05$ s.}\label{ARIMA_GPR_1}
\end{figure}
\begin{figure}
    \centering
        \includegraphics[scale=.25]{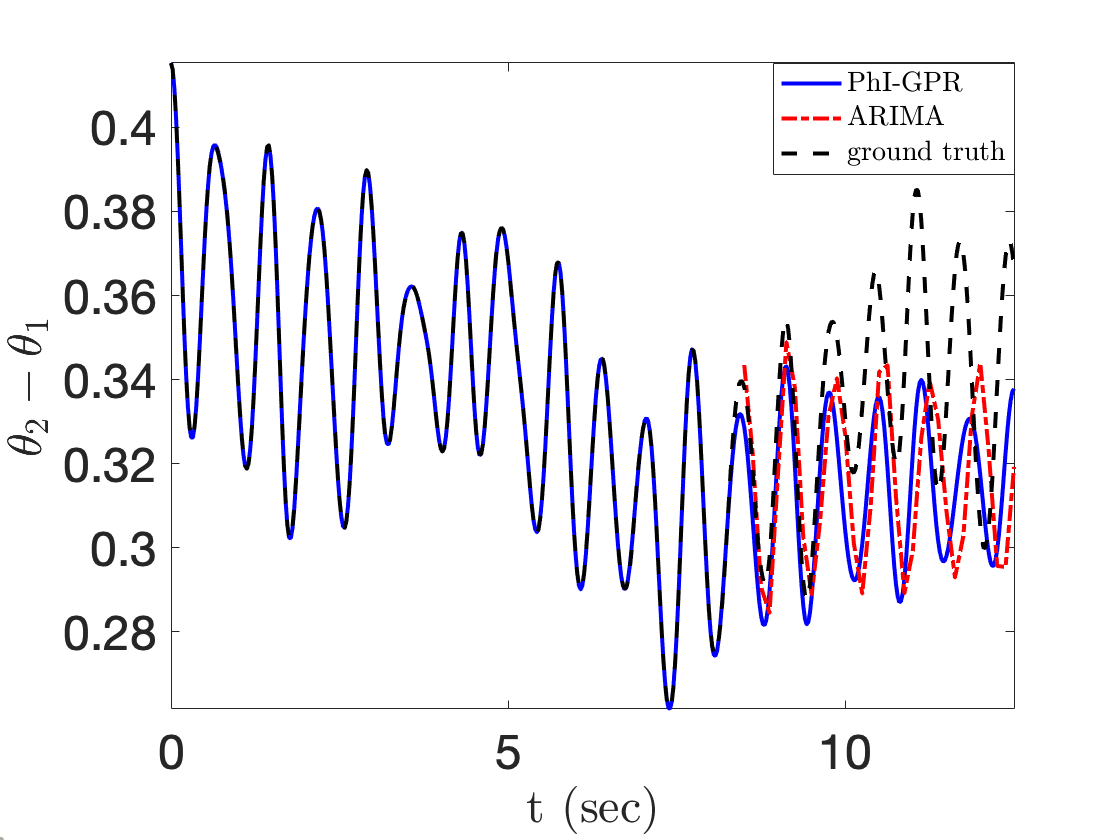}
        \caption{Forecasting of $\theta_2(t)-\theta_1(t)$ using GPR and ARIMA when measurements of $\theta_k$ and $\omega_k$ ($k=1 ,2, 3$) are available for $t<8.3375$ s every $0.125 $ s.}\label{ARIMA_GPR_2}
\end{figure}
\begin{figure}
    \centering
        \includegraphics[scale=.25]{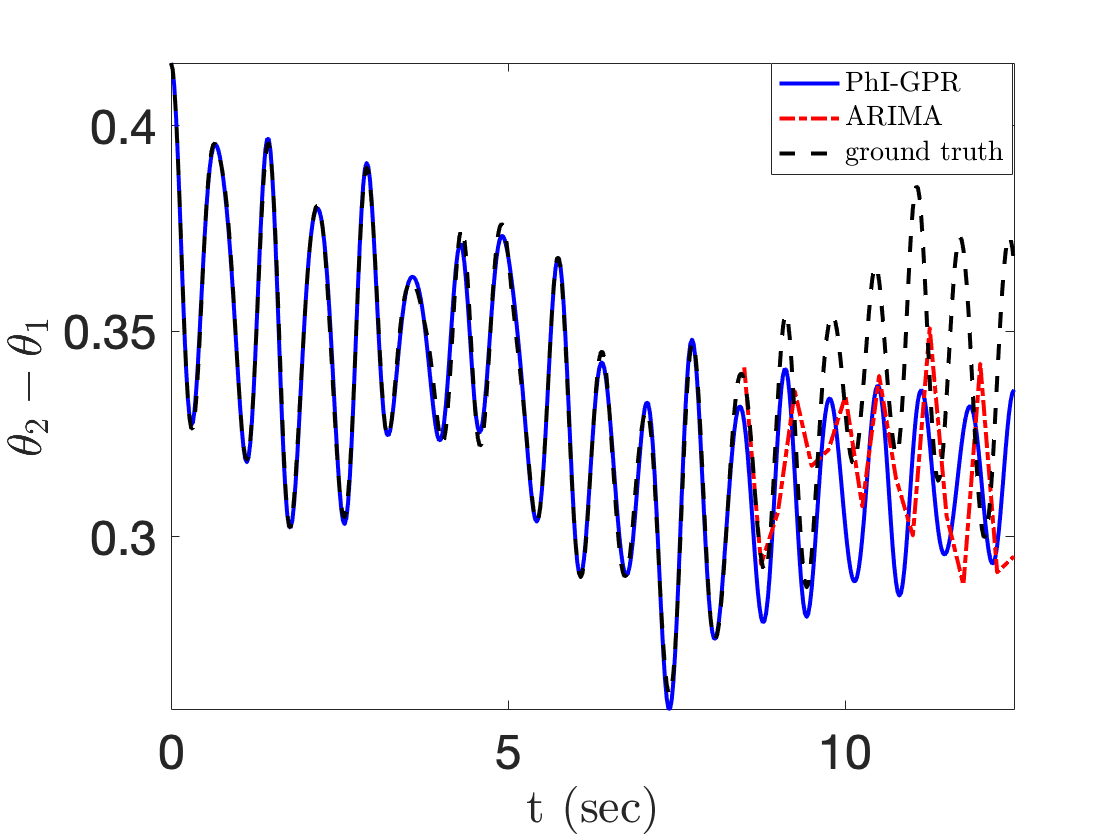}
        \caption{Forecasting of $\theta_2(t)-\theta_1(t)$ using GPR and ARIMA when measurements of $\theta_k$ and $\omega_k$ ($k=1 ,2, 3$) are available for $t<8.3375$ s every $0.25$ s.}\label{ARIMA_GPR_3}
\end{figure}
\begin{figure}
    \centering
        \includegraphics[scale=.25]{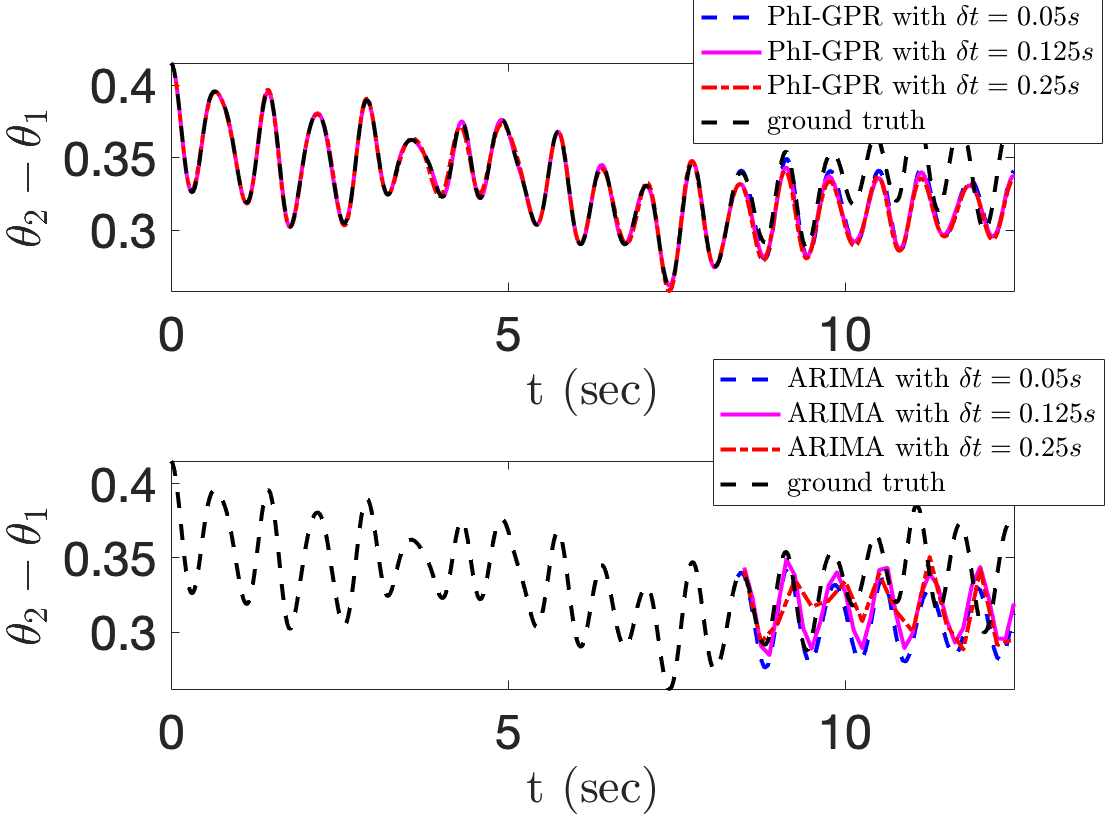}
        \caption{Forecasting of $\theta_2(t)-\theta_1(t)$ using GPR and ARIMA when measurements of $\theta_k$ and $\omega_k$ ($k=1 ,2, 3$) are available for $t<8.3375$ s.}\label{combination}
\end{figure}

As before, we assume that the measurements of $\theta_k$ and $\omega_k$ ($k=1 ,2, 3$) are available for $t<8.3375$ s, and our objective is to forecast these states for $t>8.3375$ s. We test the performance of PhI-GPR and ARIMA for three cases when the data are collected every 0.05 s, 0.125 s, and 0.25 s, respectively. In the first case (the data are collected every $0.05$ s), the ARIMA model for $\theta_2(t)-\theta_1(t)$ forecasting is ARIMA$(15,0,1)$.  PhI-GPR is also used to estimate states from the data for $t < 8.3375$ s  available every 0.025 s.  
Figure \ref{ARIMA_GPR_1} shows the PhI-GPR and  ARIMA forecasts of $\theta_2(t)-\theta_1(t)$. We can see that PhI-GPR provides a more accurate forecast than ARIMA, especially for the first two seconds. 

Next, we consider two cases when the measurements of $\theta_k$ and $\omega_k$ are available every $0.125$ s and $0.25$ s for $t<8.3375$ s, respectively. As before, we aim to forecast $\theta_2(t)-\theta_1(t)$  for $t>8.3375$ s.  
PhI-GPR is also used to estimate $\theta_2(t)-\theta_1(t)$ from the data for $t< 8.3375$ s every 0.025 s. 
In these two cases, the ARIMA model for $\theta_2(t)-\theta_1(t)$ is ARIMA$(15,0,1)$. The ARIMA and PhI-GPR forecast of $\theta_2(t)-\theta_1(t)$ as well as the ground truth are shown in figures \ref{ARIMA_GPR_2} and \ref{ARIMA_GPR_3}. As before, the PhI-GPR forecast is more accurate than ARIMA. These figures also show that the PhI-GPR estimate of $\theta_2(t)-\theta_1(t)$ is in a good agreement with the ground truth.

Figure \ref{combination}  compares ARIMA and PhI-GPR performance when data is available every $\delta t = 0.05$ s, $0.125$ s, and $0.25$ s. We can see that the ARIMA forecast is sensitive to $\delta t$, while the PhI-GPR prediction is practically independent of $\delta t$ as long as $\delta t$ is smaller than the correlation time of the states, which is approximately 2 s for the considered system. The accuracy of the ARIMA forecast increases with decreasing $\delta t$. For the smallest tested $\delta t$, we find that the accuracy of PhI-GPR is higher than ARIMA for the first two seconds and then comparable with ARIMA after two seconds. 

\subsection{The effect of measurement noises on PhI-GPR and ARIMA forecasting}\label{noisy}
Finally, we consider the effect of measurement noise on PhI-GPR and ARIMA forecasting. We study two cases where we add 1\% and 5\% noise, correspondingly, to $\omega_k$ and $\theta_k$ measurements. 
\begin{figure}
    \centering
        \includegraphics[scale=.25]{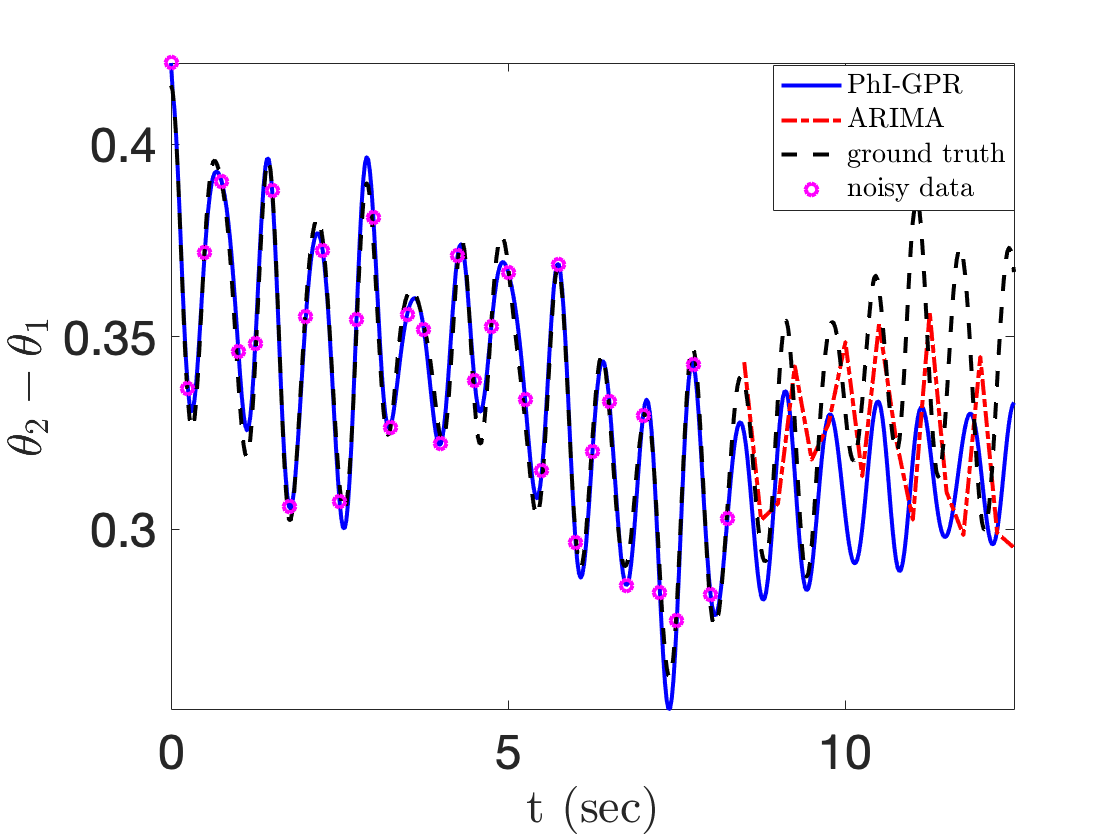}
        \caption{Forecasting of $\theta_2(t)-\theta_1(t)$ using GPR and ARIMA when measurements of $\theta_k$ and $\omega_k$ ($k=1 ,2, 3$) are available for $t<8.3375$ s every $0.25$ s with 1\% measurement noise.}\label{com_tot1}
\end{figure}
\begin{figure}
    \centering
        \includegraphics[scale=.25]{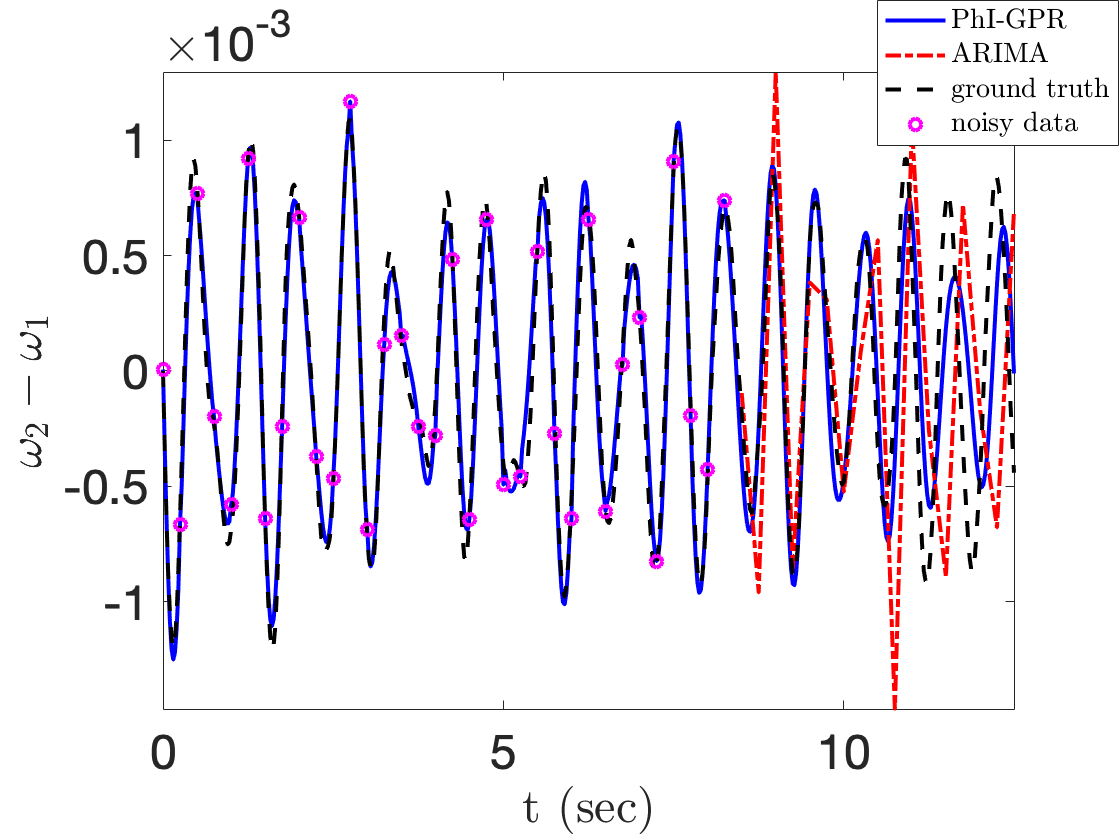}
        \caption{Forecasting of  $\omega_2(t)-\omega_1(t)$ using GPR and ARIMA when measurements of $\theta_k$ and $\omega_k$ ($k=1 ,2, 3$) are available for $t<8.3375$ s every $0.25$ s with 1\% measurement noise.}\label{com_too1}
\end{figure}
\begin{figure}
    \centering
        \includegraphics[scale=.25]{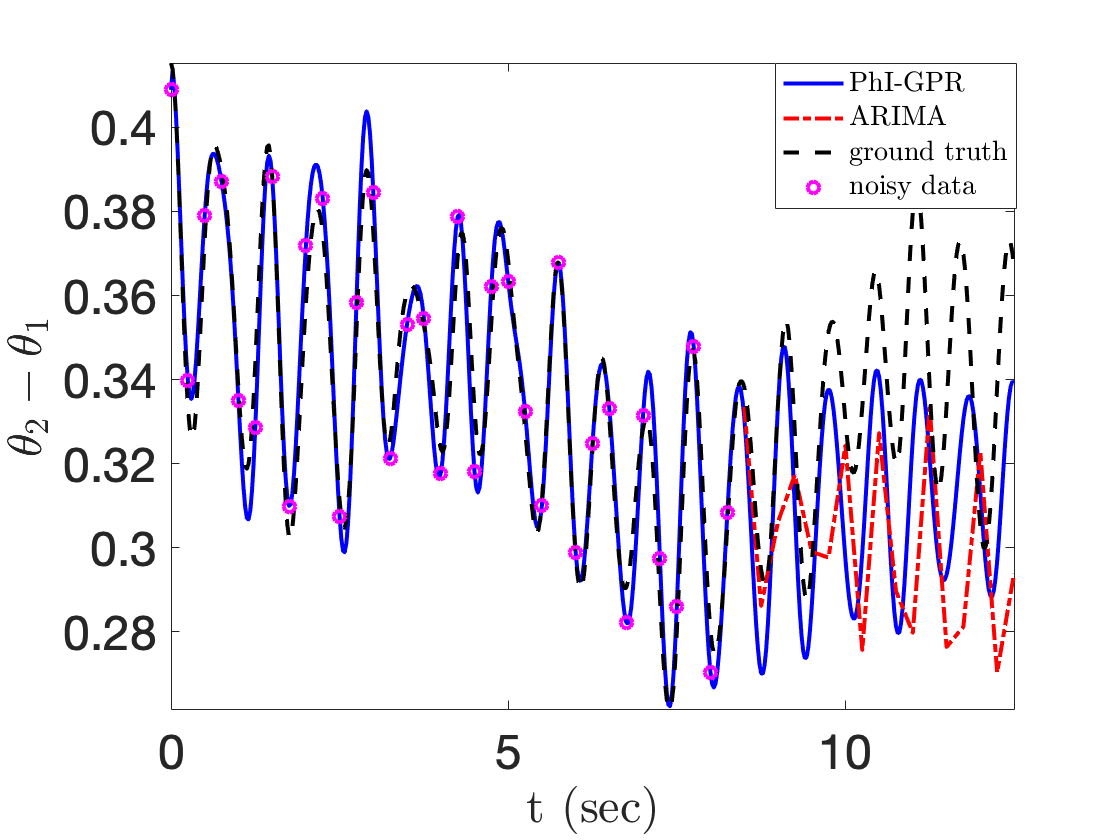}
        \caption{Forecasting of $\theta_2(t)-\theta_1(t)$ using GPR and ARIMA when measurements of $\theta_k$ and $\omega_k$ ($k=1 ,2, 3$) are available for $t<8.3375$ s every $0.25$ s with 5\% measurement noise.}\label{com_tot5}
\end{figure}
\begin{figure}
    \centering
        \includegraphics[scale=.25]{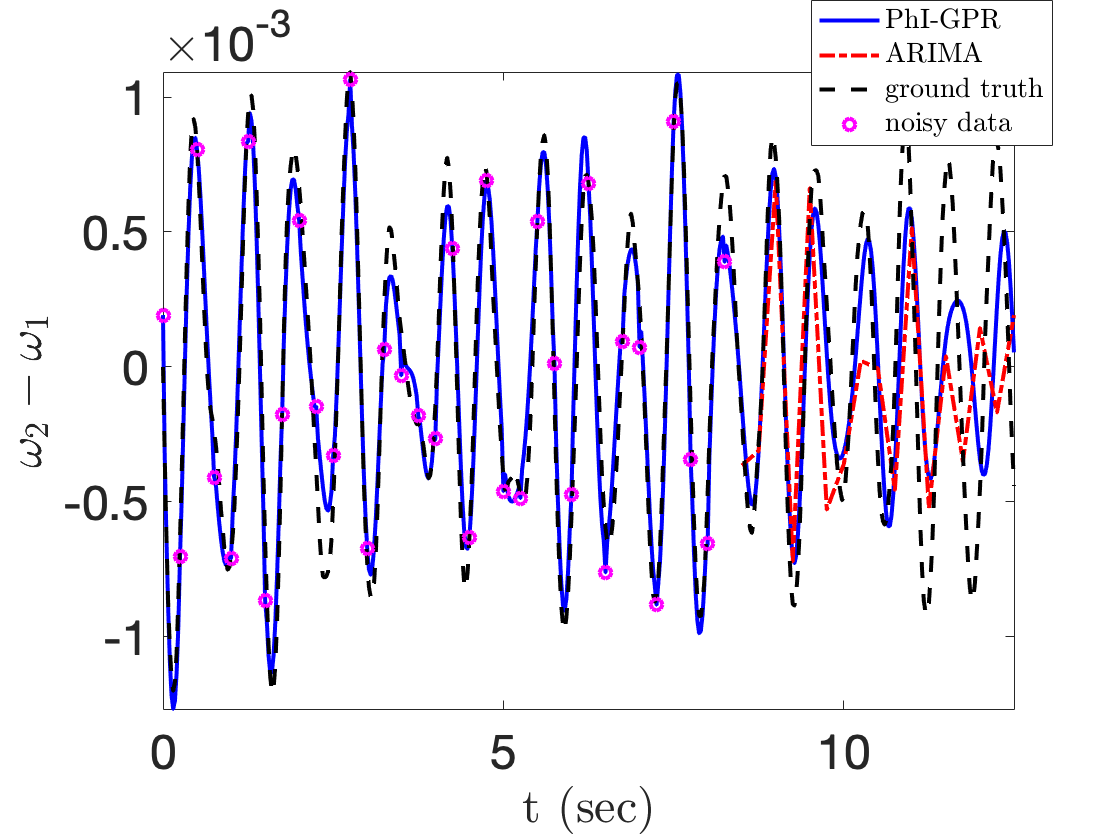}
        \caption{Forecasting of  $\omega_2(t)-\omega_1(t)$ using GPR and ARIMA when measurements of $\theta_k$ and $\omega_k$ ($k=1 ,2, 3$) are available for $t<8.3375$ s every $0.25$ s with 5\% measurement noise.}\label{com_too5}
\end{figure}

We assumed that (noisy) measurements of $\theta_k$ and $\omega_k$ are available for $t<8.3375$ s every $0.25$ s. The ARIMA models for $\theta_2(t)-\theta_1(t)$  and $\omega_2(t)-\omega_1(t)$ forecasting are  both ARIMA$(15,0,1)$.
 The PhI-GPR and ARIMA forecasts of  $\theta_2(t)-\theta_1(t)$ and $\omega_2(t)-\omega_1(t)$ based on data with 1\% and 5\% noise  are shown in 
 figures \ref{com_tot1} -- \ref{com_too1}  and \ref{com_tot5} -- \ref{com_too5}, respectively.  These figures also show the PhI-GPR states estimation every 0.025 s for $t < 8.3375$ s.  We can see that the PhI-GPR forecast is more accurate than ARIMA, especially for the first two seconds. Also, the PhI-GPR forecast is significantly less sensitive to the measurement noise than the ARIMA forecast, which significantly worsens as the noise increases. The PhI-GPR estimation of $\theta_2(t)-\theta_1(t)$ and $\omega_2(t)-\omega_1(t)$ for $t<8.3375$ s is in good agreement with the ground truth  for both noise levels. 

\section{Discussion and conclusions}
\label{sec:Conclusions}

The PhI-GPR method for short-term forecasting and state estimation of the phase angle, angular speed, and mechanical wind power of an $N$-generator power grid system with partial measurements is presented. The traditional data-driven GPR method estimates the prior mean and covariance functions from measurements by maximizing the so-called marginal likelihood function, whereas the PhI-GPR method computes covariance from  partially known swing equations describing the power grid dynamics where the unknown variables (in our case, mechanical wind power) are treated as random processes. 
Therefore, unlike data-driven GPR, the PhI-GPR method can be used to estimate and forecast even unobserved variables. For example, given observations of the angular velocity of generators, PhI-GPR is able to forecast and estimate all system states, including the angular velocity, phase angle and mechanical power of these generators.
For the considered power grid system consisting of two wind generators and one traditional generator, we find that PhI-GPR provides an accurate estimation of the unobserved states and an accurate forecast of observed and unobserved states for times smaller than the correlation time of the system. For larger times, the PhI-GPR forecast stays within two standard deviations of the ground truth. 

We also provide a comparison between PhI-GPR and ARIMA, a standard forecasting method. Like the data-driven GPR, ARIMA can only forecast observed variables. For observed variables, we demonstrate that  PhI-GPR is at least as accurate as ARIMA when the time between observations $\delta t$ is sufficiently small. As $\delta t $ increases, the accuracy of the ARIMA forecast deteriorates, while PhI-GPR remains accurate as long as $\delta t$ is smaller than the correlation time of forecasted variables.  We find that the PhI-GPR and ARIMA forecasts deteriorate in the presence of measurement noises, but for the same considered noise levels, PhI-GPR remains more accurate than ARIMA. 

In this work, we focus on forecasting short-term dynamics of power grid systems and use swing equations to model data at the scale of tens of seconds. Our method can be extended to forecast power grid systems at larger time scales, in which case the covariance can be computed from the power flow equations. 

The accuracy of the PhI-GPR method depends on the fidelity of the physics-based model that is used to compute the covariance functions. In general, this is not a problem for modeling power grids because the  equations describing the power grid behavior are well established. The main computational cost of PhI-GPR is associated with computing the covariance functions, which requires solving the governing equations multiple times (for different realizations of unknown parameters). In our future work, we will investigate the use of linearized equations, which would allow simple (deterministic) equations to be derived for covariance functions that could be solved relatively quickly.

\bibliographystyle{siamplain}

\begin{thebibliography}{10}

\bibitem{arnold1974stochastic}
{\sc L.~Arnold}, {\em Stochastic differential equations}, New York,  (1974).

\bibitem{brockwell2016introduction}
{\sc P.~J. Brockwell and R.~A. Davis}, {\em Introduction to time series and
  forecasting}, Springer, 2016.

\bibitem{catalao2011hybrid}
{\sc J.~Catalao, H.~Pousinho, and V.~Mendes}, {\em Hybrid wavelet-pso-anfis
  approach for short-term wind power forecasting in portugal}, IEEE
  Transactions on Sustainable Energy, 2 (2011), pp.~50--59.

\bibitem{charlton2014refined}
{\sc N.~Charlton and C.~Singleton}, {\em A refined parametric model for short
  term load forecasting}, International Journal of Forecasting, 30 (2014),
  pp.~364--368.

\bibitem{fan2011short}
{\sc S.~Fan and R.~J. Hyndman}, {\em Short-term load forecasting based on a
  semi-parametric additive model}, IEEE Transactions on Power Systems, 27
  (2011), pp.~134--141.

\bibitem{fentis2016short}
{\sc A.~Fentis, L.~Bahatti, M.~Mestari, M.~Tabaa, A.~Jarrou, and B.~Chouri},
  {\em Short-term pv power forecasting using support vector regression and
  local monitoring data}, in 2016 International Renewable and Sustainable
  Energy Conference (IRSEC), IEEE, 2016, pp.~1092--1097.

\bibitem{genton-2015-crosscovariance}
{\sc M.~G. Genton and W.~Kleiber}, {\em Cross-covariance functions for
  multivariate geostatistics}, Statistical Science, 30 (2015), pp.~147--163,
  \url{https://doi.org/10.1214/14-STS487}.

\bibitem{goude2013local}
{\sc Y.~Goude, R.~Nedellec, and N.~Kong}, {\em Local short and middle term
  electricity load forecasting with semi-parametric additive models}, IEEE
  transactions on smart grid, 5 (2013), pp.~440--446.

\bibitem{grant2014short}
{\sc J.~Grant, M.~Eltoukhy, and S.~Asfour}, {\em Short-term electrical peak
  demand forecasting in a large government building using artificial neural
  networks}, Energies, 7 (2014), pp.~1935--1953.

\bibitem{hong2010short}
{\sc T.~Hong}, {\em Short Term Electric Load Forecasting}, PhD thesis, North
  Carolina State University, 2012.

\bibitem{hong2014global}
{\sc T.~Hong, P.~Pinson, and S.~Fan}, {\em Global energy forecasting
  competition 2012}, 2014.

\bibitem{hong2014fuzzy}
{\sc T.~Hong and P.~Wang}, {\em Fuzzy interaction regression for short term
  load forecasting}, Fuzzy optimization and decision making, 13 (2014),
  pp.~91--103.

\bibitem{hong2013long}
{\sc T.~Hong, J.~Wilson, and J.~Xie}, {\em Long term probabilistic load
  forecasting and normalization with hourly information}, IEEE Transactions on
  Smart Grid, 5 (2013), pp.~456--462.

\bibitem{huang2012state}
{\sc Y.-F. Huang, S.~Werner, J.~Huang, N.~Kashyap, and V.~Gupta}, {\em State
  estimation in electric power grids: Meeting new challenges presented by the
  requirements of the future grid}, IEEE Signal Processing Magazine, 29 (2012),
  pp.~33--43.

\bibitem{hyndman2018forecasting}
{\sc R.~J. Hyndman and G.~Athanasopoulos}, {\em Forecasting: principles and
  practice}, OTexts, 2018.

\bibitem{hyndman2009density}
{\sc R.~J. Hyndman and S.~Fan}, {\em Density forecasting for long-term peak
  electricity demand}, IEEE Transactions on Power Systems, 25 (2009),
  pp.~1142--1153.

\bibitem{kusiak2009short}
{\sc A.~Kusiak, H.~Zheng, and Z.~Song}, {\em Short-term prediction of wind farm
  power: A data mining approach}, IEEE Transactions on energy conversion, 24
  (2009), pp.~125--136.

\bibitem{liu2012short}
{\sc Y.~Liu, J.~Shi, Y.~Yang, and W.-J. Lee}, {\em Short-term wind-power
  prediction based on wavelet transform--support vector machine and
  statistic-characteristics analysis}, IEEE Transactions on Industry
  Applications, 48 (2012), pp.~1136--1141.

\bibitem{lloyd2014gefcom2012}
{\sc J.~R. Lloyd}, {\em Gefcom2012 hierarchical load forecasting: Gradient
  boosting machines and gaussian processes}, International Journal of
  Forecasting, 30 (2014), pp.~369--374.

\bibitem{milshtein1994numerical}
{\sc G.~Milshtein and M.~Tret'yakov}, {\em Numerical solution of differential
  equations with colored noise}, Journal of Statistical Physics, 77 (1994),
  pp.~691--715.

\bibitem{nedellec2014gefcom2012}
{\sc R.~Nedellec, J.~Cugliari, and Y.~Goude}, {\em Gefcom2012: Electric load
  forecasting and backcasting with semi-parametric models}, International
  Journal of forecasting, 30 (2014), pp.~375--381.

\bibitem{nishikawa2015comparative}
{\sc T.~Nishikawa and A.~E. Motter}, {\em Comparative analysis of existing
  models for power-grid synchronization}, New Journal of Physics, 17 (2015),
  p.~015012, \url{https://doi.org/10.1088/1367-2630/17/1/015012},
  \url{https://doi.org/10.1088%2F1367-2630%2F17%2F1%2F015012}.

\bibitem{quinonero2005unifying}
{\sc J.~Qui{\~n}onero-Candela and C.~E. Rasmussen}, {\em A unifying view of
  sparse approximate gaussian process regression}, Journal of Machine Learning
  Research, 6 (2005), pp.~1939--1959.

\bibitem{rosenthal2018ensemble}
{\sc W.~S. Rosenthal, A.~M. Tartakovsky, and Z.~Huang}, {\em Ensemble kalman
  filter for dynamic state estimation of power grids stochastically driven by
  time-correlated mechanical input power}, IEEE Transactions on Power Systems,
  33 (2018), pp.~3701--3710.

\bibitem{snelson2006sparse}
{\sc E.~Snelson and Z.~Ghahramani}, {\em Sparse gaussian processes using
  pseudo-inputs}, in Advances in neural information processing systems, 2006,
  pp.~1257--1264.

\bibitem{song2005short}
{\sc K.-B. Song, Y.-S. Baek, D.~H. Hong, and G.~Jang}, {\em Short-term load
  forecasting for the holidays using fuzzy linear regression method}, IEEE
  transactions on power systems, 20 (2005), pp.~96--101.

\bibitem{taieb2014gradient}
{\sc S.~B. Taieb and R.~J. Hyndman}, {\em A gradient boosting approach to the
  kaggle load forecasting competition}, International journal of forecasting,
  30 (2014), pp.~382--394.

\bibitem{tartakovsky2019physicsHICSS}
{\sc A.~Tartakovsky and R.~Tipireddy}, {\em Physics-informed machine learning
  method for forecasting and uncertainty quantification of partially observed
  and unobserved states in power grids}, in Proceedings of the 52nd Hawaii
  International Conference on System Sciences, 2019.

\bibitem{thiyagarajan2016real}
{\sc K.~Thiyagarajan and R.~S. Kumar}, {\em Real time energy management and
  load forecasting in smart grid using compactrio}, Procedia Computer Science,
  85 (2016), pp.~656--661.

\bibitem{wang2015probabilistic}
{\sc P.~Wang, D.~A. Barajas-Solano, E.~Constantinescu, S.~Abhyankar, D.~Ghosh,
  B.~Smith, Z.~Huang, and A.~M. Tartakovsky}, {\em Probabilistic density
  function method for stochastic odes of power systems with uncertain power
  input}, SIAM/ASA Journal on Uncertainty Quantification, 3 (2015),
  pp.~873--896.

\bibitem{wang2016electric}
{\sc P.~Wang, B.~Liu, and T.~Hong}, {\em Electric load forecasting with recency
  effect: A big data approach}, International Journal of Forecasting, 32
  (2016), pp.~585--597.

\bibitem{williams2006gaussian}
{\sc C.~K. Williams and C.~E. Rasmussen}, {\em Gaussian processes for machine
  learning}, vol.~2, MIT press Cambridge, MA, 2006.

\bibitem{yeung2017learning}
{\sc E.~Yeung, S.~Kundu, and N.~Hodas}, {\em Learning deep neural network
  representations for koopman operators of nonlinear dynamical systems}, arXiv
  preprint arXiv:1708.06850,  (2017).

\bibitem{yoder2014short}
{\sc M.~Yoder, A.~S. Hering, W.~C. Navidi, and K.~Larson}, {\em Short-term
  forecasting of categorical changes in wind power with markov chain models},
  Wind energy, 17 (2014), pp.~1425--1439.

\end{thebibliography}

\end{document}